%% file: paper.tex
\documentclass[lettersize,journal]{IEEEtran}
\usepackage{amsmath,amsfonts}
\usepackage{algorithmic}
\usepackage{algorithm}
\usepackage{array}
\usepackage[caption=false,font=normalsize,labelfont=sf,textfont=sf]{subfig}
\usepackage{textcomp}
\usepackage{stfloats}
\usepackage{url}
\usepackage{verbatim}
\usepackage{graphicx}
\usepackage{soul}
\usepackage{tabularray}
\usepackage{makecell}
\usepackage{enumerate}
\usepackage{color,soul}
\usepackage{orcidlink}
\usepackage{booktabs}
\usepackage{amssymb}
\usepackage{xcolor} 
\usepackage[normalem]{ulem}
\usepackage[switch]{lineno}

\usepackage[
    backend=biber,
    style=ieee,
    natbib=true,
    dashed=false,      
    maxbibnames=6,     
    minbibnames=1      
]{biblatex}
\usepackage{acronym}

\acrodef{ML}{machine learning}
\acrodef{DL}{deep learning}
\acrodef{RS}{remote sensing}
\acrodef{UAV}{unmanned aerial vehicle}
\acrodef{CNN}{convolutional neural network}
\acrodef{DEM}{digital elevation model}
\acrodef{RF}{random forest}
\acrodef{SVM}{support vector machine}
\acrodef{FT}{forest type}
\acrodef{TS}{tree species}
\acrodef{LT}{leaf type}
\acrodef{OA}{overall accuracy}
\acrodef{VNIR}{visible and near-infrared}
\acrodef{SWIR}{short-wave infrared}

\begin{document}

\title{Fusing Multimodal CNN Features with Structurally-Masked Transformers for Hierarchical Sub-Pixel Tree Species Mapping}

\author{Mattia Ferrari\orcidlink{0009-0008-0864-5344}\thanks{This work was partially supported by Italian Space Agency (ASI) under Contract 2022-1-U.0, by the project “AFORISMA”.

Mattia Ferrari and Lorenzo Bruzzone are with the Department of Information Engineering and Computer Science, University of Trento, 38123 Trento, Italy (e-mail: mattia.ferrari-2@unitn.it; lorenzo.bruzzone@unitn.it).

\par\medskip
\noindent
This work has been submitted to the IEEE for possible publication.
Copyright may be transferred without notice, after which this
version may no longer be accessible.}, and Lorenzo Bruzzone\orcidlink{0000-0002-6036-459X},~\IEEEmembership{Fellow,~IEEE}}

\markboth{Under review for publication on IEEE Transactions on Geoscience and Remote Sensing}
{Ferrari and Bruzzone: Fusing Multimodal Features for Sub-Pixel Tree Species Mapping}

\maketitle

\begin{abstract}
Precise forest species mapping using spaceborne hyperspectral data remains a significant challenge due to high spectral similarity and the inherent sub-pixel heterogeneity of natural forest stands. While traditional deep learning models exploit spatial context through convolutional architectures, such approaches often introduce noise in fragmented or mixed-species forest environments. In this paper, we propose a hierarchy-aware, multimodal deep learning framework designed to resolve sub-pixel forest composition by integrating PRISMA (PRecursore IperSpettrale della Missione Applicativa) hyperspectral imagery, high-resolution panchromatic data, and digital elevation models. The architecture features a multi-branch encoder—leveraging an EfficientNetV2-based one-dimensional convolutional neural network (CNN) for spectral signatures and twin two-dimensional CNNs for spatial and topographic features—coupled with a novel structurally-masked Transformer. By using a binary transition tensor to condition the Transformer slot embeddings and constrain inference, the model enforces taxonomic consistency between broad forest categories and fine-grained species distributions. The framework is validated across three diverse Italian biogeographical regions (the Alps and the Central and Southern Apennines) using a unique dataset of field-surveyed plots. Experimental results demonstrate that the proposed method consistently outperforms state-of-the-art multi-label and ranking-based paradigms, particularly in identifying subordinate species and their relative dominance. Notably, the bidirectional learning mechanism allows sub-pixel compositional features to refine macroscopic forest type predictions, while the integration of topographic priors effectively resolves spectral ambiguities in complex terrains. The proposed approach provides a knowledge-driven approach to regional-scale biodiversity monitoring and forest characterization within the evaluated PRISMA scenarios.
\end{abstract}


\section{Introduction}
\input{chapters/introduction_short}

\section{Related Work}
\label{sec:rw}
\input{chapters/related_works}

\section{Proposed Methodology}
\label{sec:pm}
\input{chapters/methodology}

\section{Datasets Description}
\label{sec:dd}
\input{chapters/dataset}

\section{Experimental Setup}
\label{sec:es}
\input{chapters/setup}

\section{Experimental Results}
\label{sec:er}
\input{chapters/results_ft}
\input{chapters/results_ts}
\input{chapters/results_ablation}

\section{Conclusions}
\label{sec:co}
\input{chapters/conclusions}

\printbibliography

\end{document}

%% file: chapters/introduction_short.tex
\IEEEPARstart{F}{orests} are among the most complex and indispensable ecosystems on Earth, providing critical ecological services such as carbon sequestration, biodiversity conservation, hydrological regulation, and climate moderation. Increasing pressures from deforestation, land-use change, and climate variability underline the need for monitoring frameworks that are accurate, consistent, and scalable.

Within this context, forest classification provides essential baseline data for ecosystem modeling, conservation planning, and evidence-based environmental policy. Classification systems can be organized according to different criteria. Physiognomic classifications rely on structural characteristics such as broadleaf versus needleleaf or evergreen versus deciduous, and are widely adopted in regional and global land-cover assessments. Floristic classifications can be implemented at varying levels of detail. Coarser classifications group forests into categories based on species assemblages (\ac{FT} classification). In contrast, species-level mapping (\ac{TS} classification) identifies individual species distributions, providing detailed insights into biodiversity and ecological processes. While both rely on species composition, their thematic resolution and objectives differ: the former generalizes forest communities into recognizable types, whereas the latter captures fine-grained ecological variation. However, species-level mapping remains considerably more challenging due to high species richness, spectral similarity among species, and the prevalence of mixed-species canopies in many biomes. These factors amplify spectral mixing within the spatial resolution of multispectral and hyperspectral sensors, thereby complicating classification tasks \cite{upadhyay2018hyperspectral}.

Traditional forest classification relied on ground surveys and aerial photointerpretation, which, while accurate, are costly, time-consuming, and limited in coverage. \ac{RS} has fundamentally changed this paradigm. Multispectral satellite missions such as Landsat and Sentinel-2 have long been employed for vegetation monitoring thanks to their sensitivity to chlorophyll, leaf water content, and canopy structure. However, their relatively coarse spectral resolution restricts their ability to separate spectrally similar species. In contrast, airborne and \ac{UAV}-based hyperspectral sensors, offering hundreds of narrow bands, provide the detail required for species-level discrimination but remain constrained by limited coverage and irregular acquisition. The advent of spaceborne hyperspectral missions such as PRISMA (PRecursore IperSpettrale della Missione Applicativa) \cite{candela2016prisma} and EnMAP (Environmental Mapping and Analysis Program) \cite{guanter2015enmap} represents a significant advancement, integrating high spectral resolution with moderate spatial coverage. While PRISMA and EnMAP sensors provide a coarser spatial resolution than Sentinel-2, the contiguous spectral sampling allows the detection of diagnostic absorption features associated with pigments, water content, and structural biochemistry, enabling more accurate forest characterization \cite{vangi2021new}.

\Acf{ML} techniques have been extensively employed for forest classification; however, traditional approaches typically rely on feature engineering. Although feature extraction and feature selection can enhance classification performance, these processes often require expert knowledge and may inadequately represent the nonlinear and high-dimensional relationships inherent in hyperspectral data. In contrast, \ac{DL} approaches, particularly \acp{CNN}, recurrent neural networks, and, more recently, Transformer-based architectures, are capable of automatically learning meaningful features directly from raw data \cite{paoletti2019deep, hong2022spectralformer}. By modeling spectral and spatial dependencies, \ac{DL} methods mitigate the need for manual feature engineering and have demonstrated superior performance in forest mapping tasks.

\ac{TS} classification poses unique challenges compared to broader \ac{FT} classification, largely depending on the spatial structure of the target ecosystem. While certain biomes, such as boreal forests, often consist of relatively homogeneous stands, many temperate and complex natural environments, such as alpine forests, exhibit a high degree of species intermixing \cite{kluczek2023mountain}. In these heterogeneous scenarios, individual trees of different species frequently coexist within the same stand, resulting in strong sub-pixel heterogeneity and irregular spatial distributions. Under these conditions, the use of spatial-contextual features commonly exploited in \ac{DL} for \ac{RS} applications becomes less effective. In fact, incorporating spatial context may introduce additional noise rather than improve discrimination, as neighboring pixels do not necessarily belong to the same class in complex or highly heterogeneous environments \cite{wan2020hyperspectral}. Consequently, in highly intermixed environments, models for species-level forest classification must rely more heavily on spectral signatures at the pixel level or on very small spatial neighborhoods, increasing the demand for precise and robust spectral feature extraction.

Nevertheless, ancillary information can be exploited to complement hyperspectral observations and potentially alleviate some of these challenges. For instance, the high-resolution panchromatic band provided by hyperspectral missions such as PRISMA can contribute additional spatial detail to the spectral information of hyperspectral pixels, which may help characterize canopy structure and spatial heterogeneity \cite{perretta2024testing}. Similarly, \acp{DEM} can provide complementary topographic information, as variables such as elevation, slope, and aspect influence species distributions through their effects on microclimate, soil properties, and water availability \cite{vorovencii2024assessing}.

Traditional spectral unmixing approaches \cite{bioucas2012hyperspectral} are commonly employed to address sub-pixel heterogeneity by estimating the continuous fractional abundances of distinct materials within a pixel. However, these techniques face fundamental limitations when applied to species-level mapping in complex forest ecosystems. High intra-species spectral variability, driven by leaf age, phenological stage, illumination geometry, and canopy structure, often exceeds inter-species differences, fundamentally undermining the extraction of stable endmembers. Additionally, conventional unmixing typically assumes a simplified physical mixing process, which fails to adequately capture the complex, nonlinear interactions occurring within multilayered forest canopies. Consequently, while unmixing is valuable for estimating broad vegetation fractions, it becomes unreliable for fine-grained \ac{TS} discrimination. Instead of estimating continuous fractional abundances via pure endmembers, our proposed methodology frames sub-pixel heterogeneity as a multi-label compositional problem. By leveraging a \ac{DL} framework, our model learns complex, nonlinear spectral features directly from mixed ground-truth data, entirely bypassing the stringent prerequisites of endmember extraction and simplified mixing models.

Another critical bottleneck concerns the availability of ground-truth data. While \ac{FT} labels can be efficiently derived from photointerpretation or existing forest inventories, resulting in larger and more balanced datasets, collecting reliable \ac{TS} samples is considerably more challenging. Field data acquisition is labor-intensive, spatially biased toward accessible areas, and often insufficient to capture the variability of heterogeneous landscapes. This imbalance between category- and species-level annotations significantly constrains fine-grained classification.

To address these limitations, this paper proposes a hybrid \ac{DL} architecture. In our approach, one-dimensional (1D) \acp{CNN} are utilized to exploit the sequential nature of spectral signatures \cite{hu2015deep}, acting as robust, lightweight extractors that are significantly less data-demanding than pure, large-scale Transformer backbones. We further enhance this framework through a multi-branch design: while the main 1D \ac{CNN} processes the hyperspectral signature, additional compact two-dimensional (2D) \ac{CNN} branches are integrated into our model to handle auxiliary spatial information: PRISMA panchromatic data and topographic inputs (\ac{DEM}). To explicitly model the ecological hierarchical relationship between \ac{FT} and \ac{TS}, we introduce a novel bidirectional relational mechanism. Our methodology posits that since \acp{FT} can be classified more reliably, they can constrain and guide the \ac{TS} classification \textit{top-down}. Conversely, our model leverages sub-pixel species composition to resolve ambiguities at the \ac{FT} level \textit{bottom-up}. By embedding this dependency within a structured attention mechanism, the proposed architecture reduces classification uncertainty and ensures semantic consistency across taxonomic levels.

Building upon these principles, one of the main novelties of this study lies in how this hierarchy-aware framework jointly exploits hyperspectral imagery, panchromatic data, and \acp{DEM} for sub-pixel forest classification. Rather than predicting a single dominant label per pixel or attempting to estimate continuous fractions via unstable unmixing algorithms, our approach explicitly predicts the discrete composition of up to three co-occurring species. The proposed method operates in two stages. First, a multi-branch \ac{CNN} encoder is pretrained to extract robust baseline features and classify broad \acp{FT}, leveraging abundant photo-interpreted labels. In the second stage, the \ac{CNN} weights are frozen, and the extracted features are fed into a novel structurally-masked Transformer. This module is trained on the scarce \ac{TS} ground-truth data to perform the joint \ac{FT}-\ac{TS} classification. By using the ecological transition tensor to condition the Transformer slot embeddings and constrain inference, the framework enforces strict structural constraints, guiding the network to ecologically consistent sub-pixel compositions.

The key contributions of this work can be summarized as follows:
\begin{enumerate}
    \item \textbf{Multi-Branch Architecture for Multimodal Fusion:} We propose a multi-branch architecture capable of fusing spaceborne hyperspectral data (PRISMA) with panchromatic imagery and topographic variables (\ac{DEM}), enabling accurate forest classification over heterogeneous landscapes.
    \item \textbf{Sub-Pixel Cascaded Classification:} To address the severe spatial intermixing typical of complex natural forest ecosystems, we introduce a \ac{TS} cascaded prediction strategy that resolves sub-pixel heterogeneity in complex forest stands. The method identifies up to three species per pixel through a hierarchical dependence, where the classification of primary species guides the prediction of subordinate ones.
    \item \textbf{Structurally-Masked Hierarchical Learning:} We design a custom joint \ac{FT}-\ac{TS} Transformer that combines topology-constrained self-attention with formal ecological constraints, derived from forestry inventories, through hierarchy-aware conditioning and inference. This bidirectional, hierarchy-aware modeling supports information exchange between broad \ac{FT} and fine-grained species compositions; when the predefined constraints are used, deterministic inference ensures consistency with the encoded hierarchy.
\end{enumerate}

The remainder of this paper is organized as follows. Section \ref{sec:rw} reviews related work on forest classification. Section \ref{sec:pm} details the proposed approach. The datasets are described in Section \ref{sec:dd}, followed by Section \ref{sec:es}, which presents the experimental setup. Section \ref{sec:er} discusses the results, and finally, Section \ref{sec:co} concludes the paper and outlines future research directions.

%% file: chapters/related_works.tex
\subsection{\Acl{ML} in Hyperspectral Forest Classification}
Several studies have investigated forest classification using hyperspectral data and \ac{ML}, mostly relying on airborne acquisitions \cite{fassnacht2016review}. These studies cover diverse geographical regions and forest ecosystems, including tropical rainforests \cite{clark2005hyperspectral}, African savannas \cite{cho2012mapping}, and boreal forests \cite{dalponte2012tree}. The literature consistently demonstrates that traditional \ac{ML} algorithms, such as \ac{SVM} or \ac{RF}, can achieve high classification accuracy at the \ac{TS} level, provided that reliable ground-truth data and effective dimensionality reduction strategies are employed \cite{clark2005hyperspectral,dalponte2012tree,feret2012tree,ballanti2016tree}. However, handcrafted features and dimensionality reduction techniques strongly influence species separability, leading to substantial variability in classification outcomes depending on the chosen feature engineering pipeline \cite{dalponte2012tree,fassnacht2014comparison,hycza2018potential}.

Building on these limitations, research has increasingly applied \ac{DL} methods to hyperspectral forest classification, bypassing manual feature engineering by learning discriminative representations directly from raw spectral data \cite{paoletti2019deep}. Early \ac{DL} studies utilized 1D \acp{CNN} to effectively capture the continuous sequential spectral signatures of \acp{TS}. By treating spectral bands as a 1D sequence, these architectures analyze the pure spectral profile of each pixel independently. For instance, Xi et al. \cite{xi2019mapping} demonstrated that a 1D-CNN could extract deep feature vectors from hyperspectral profiles, outperforming traditional \ac{RF} classifiers. This was particularly evident in distinguishing broadleaf species characterized by highly overlapping spectral reflectance, using Orbita Hyperspectral Satellite (OHS-1) data. Similarly, Forgáč et al. \cite{forgavc2023contribution} combined a pixel-based 1D-CNN with principal component analysis to classify \acp{TS} from aerial hyperspectral data, leveraging 1D convolutions to extract features from spectral sequences.

Subsequent research explored 2D and three-dimensional (3D) \ac{CNN} architectures aimed at integrating spatial context alongside spectral information \cite{polonen2018tree,fricker2019convolutional,zhang2020three,nezami2020tree,sothe2020evaluating}. To process spatial-spectral features, these models inherently require the extraction of local spatial windows around each target pixel. As highlighted by a comprehensive review \cite{zhong2024review}, determining the optimal patch size is a critical and highly sensitive step in \ac{CNN}-based vegetation mapping. While integrating a spatial neighborhood in high-resolution hyperspectral data (typically captured by UAVs) effectively captures structural canopy features in broad, homogeneous stands, it becomes problematic in dense, heterogeneous forest environments. In such mixed scenarios, a fixed spatial window inevitably incorporates the spectral signatures of adjacent, differing \acp{TS}. This issue is further exacerbated when dealing with satellite hyperspectral data, where intra-pixel mixing of \acp{TS} occurs. Recent studies exploring the limits of 2D and 3D \acp{CNN} in complex forest scenes confirm that this forced spatial aggregation introduces severe spectral mixing and contextual noise \cite{chaity2024exploring}. Consequently, standard 3D architectures often degrade classification accuracy in situations where the pure spectral profile of the central pixel alone could be sufficiently discriminative \cite{fassnacht2016review}.

Beyond traditional architectures, Transformer-based models were introduced to capture global dependencies in hyperspectral data \cite{hong2022spectralformer}; however, their inherent computational inefficiency suggests they are most effective when integrated with CNNs to balance local and global feature extraction. While Vision Transformers (ViTs) may show potential for high-resolution UAV-acquired imagery, their application to satellite hyperspectral data remains suboptimal. Similar to the challenges faced by 2D and 3D CNNs in heterogeneous environments, ViTs suffer from excessive spatial mixing, compounded by a significant demand for labeled samples. More efficient alternatives like Mamba \cite{pan2024hyperspectral} have recently emerged, leveraging state-space models to achieve linear complexity; yet, given the typical scale of \acp{TS} datasets, 1D-CNNs generally remain the superior choice for data efficiency. Furthermore, while Hyperspectral Foundation Models \cite{braham2025spectralearth, wang2025hypersigma} enable robust representation learning via massive-scale pretraining, their adaptation for \ac{TS} classification is often hindered by the computationally demanding and data-intensive nature of fine-tuning large-scale architectures. Despite these rapid paradigm shifts in broader \ac{RS}, the adoption and validation of such advanced technologies specifically for \ac{TS} classification remain limited in the current literature.

At larger spatial scales, research has historically relied on multispectral data, which struggle to discriminate spectrally similar \acp{TS} \cite{hoscilo2019mapping,breidenbach2021national}. The recent availability of spaceborne hyperspectral data (e.g., PRISMA \cite{vangi2021new}, EnMAP \cite{vanguri2024mapping}) has substantially enhanced our ability to characterize vegetation dynamics at regional scales. For instance, a study mapped forest habitats using PRISMA data and \ac{RF}, significantly outperforming Sentinel-2 \cite{laurin2025monitoring}. Similarly, another study demonstrated the superiority of \ac{CNN}s over traditional algorithms for broad classification tasks using hyperspectral satellite imagery \cite{delogu2024leveraging}.

Despite these advancements, mapping heterogeneous forests at a 30-m spatial resolution introduces severe spectral mixing challenges \cite{somers2014tree}. At this scale, a single pixel covers approximately 900 m$^2$, an area that, in complex forest ecosystems, rarely corresponds to a single species. Traditionally, sub-pixel analysis has relied on Spectral Mixture Analysis (SMA) or Multiple Endmember Spectral Mixture Analysis (MESMA). However, the application of supervised hyperspectral unmixing to \ac{TS} mapping remains sparse in the literature for two primary reasons. First, the complex multiple scattering occurring within the forest canopy often violates the linear mixing assumptions required by standard SMA, leading to significant estimation errors. Furthermore, the intra-class variability characteristic of forest environments makes the extraction of ``pure'' taxonomic endmembers a highly ill-posed problem \cite{somers2014tree, stagakis2016estimating}.

As an alternative to explicit unmixing, multi-label classification offers a promising framework for identifying multiple co-occurring \acp{TS} within a single observation \cite{karalas2016land, sumbul2020deep}. By bypassing the need for endmember extraction, this approach can effectively account for the spectral complexity of mixed pixels. Nevertheless, its application to \ac{TS} classification remains largely unaddressed; current research predominantly operates under the single-label paradigm, focusing on identifying the dominant species per pixel and thereby failing to capture the inherent taxonomic diversity of heterogeneous forest stands. Moreover, standard multi-label frameworks typically produce an unstructured set of predicted classes, identifying presence but failing to quantify the relative importance or dominance of each species. To overcome these limitations, our work introduces a cascaded prediction strategy. The cascade provides an explicit formulation for representing the ordinal structure of dominant and secondary species within each pixel. This approach explicitly resolves the sub-pixel composition by establishing a strict ranking of dominant and secondary species, effectively bridging the gap between discrete classification and full spectral unmixing.

\subsection{Multimodal Data Fusion and Contextual Information}
A critical advancement in forest classification lies in the integration of heterogeneous data sources. While hyperspectral imagery provides exceptional spectral detail, its moderate spatial resolution limits structural analysis. To mitigate this, complementary high-resolution panchromatic data, natively provided by systems such as PRISMA, can provide additional spatial information related to canopy boundaries and local spatial textures \cite{vanguri2024mapping}. In parallel, structural and topographic data contribute distinct forms of information. While airborne light detection and ranging (LiDAR) provides accurate 3D reconstructions of the canopy \cite{dalponte2008fusion,cho2012mapping}, its availability is spatially limited. Conversely, spaceborne-acquired \ac{DEM}s represent a globally available topographic context, capturing elevation, slope, and aspect, which indirectly influence species distribution through microclimatic and edaphic constraints.

Recent methodologies in forestry mapping have increasingly emphasized the fusion of high-dimensional inputs to exploit complementary spectral, spatial, and structural information. For instance, the integration of UAV-borne LiDAR and hyperspectral imaging has been shown to facilitate deep feature extraction for individual \ac{TS} discrimination \cite{quan2023tree}. To fully exploit such multimodal datasets, advanced \ac{DL} architectures have been widely adopted. Ma et al. \cite{ma2024deep} demonstrated the effectiveness of integrating 1D, 2D, and 3D \acp{CNN} coupled with attention mechanisms to process combined hyperspectral and LiDAR data. Similarly, Yu et al. \cite{yu2025forest} introduced a deep ensemble learning framework designed to capture spatiotemporal-spectral features by fusing multitemporal, high-resolution spatial, and hyperspectral data.

Despite the efficacy of these multimodal models, their complexity often necessitates large annotated datasets. To address this, multi-branch architectures have emerged as a highly efficient strategy for handling heterogeneous inputs, particularly in data-limited scenarios. In such frameworks, separate branches process different modalities independently before feature fusion \cite{zhang2017spectral, ma2018hyperspectral}. Following this paradigm, our proposed framework utilizes a 1D branch dedicated to spectral signatures and separate 2D branches for panchromatic and \ac{DEM} data. This architecture allows complementary spatial and topographic information to be incorporated while maintaining a compact structure.

\subsection{Hierarchical Approaches}
Hierarchical modeling exploits the inherent nested taxonomic structure of forest ecosystems. Broader \ac{FT} categories are generally easier to classify due to higher spectral separability and the availability of larger datasets. Conversely, fine-grained \ac{TS} classification is constrained by high spectral similarity and label scarcity. In \ac{RS}, hierarchical classification has conventionally been addressed using decision trees or cascaded \ac{ML} systems \cite{chen2020mapping}. Recently, \ac{DL} architectures have been specifically adapted to explicitly incorporate these taxonomic and semantic relationships. In the broader context of \ac{RS}, hierarchical \ac{CNN}s have proven effective in capturing general discriminant features at coarse levels, while dedicating specialized learning capacity to harder-to-distinguish fine-grained classes \cite{sen2022hierarchical}. However, traditional cascaded hierarchical structures remain prone to the accumulation of classification errors across successive levels \cite{sen2022hierarchical}. Addressing this challenge within the forestry domain, Liu et al. \cite{liu2026switcher} proposed Switcher-HNet, a switchable hierarchical convolutional network designed to embed a biological taxonomy that progresses from foliage type down to specific \acp{TS}. By internalizing these relationships, the model captures shared traits among related species to prevent taxonomically inconsistent predictions, such as misclassifying a broadleaf tree as a needleleaf species. However, the taxonomic gap between broad foliage categories and individual species is considerable, potentially overlooking critical intermediate variables. A transition from \ac{FT} to \ac{TS} may provide a more ecologically informative intermediate representation; by incorporating the ecological community and environmental context inherent in \acp{FT}, the model can better constrain the search space and leverage the associative patterns that naturally define species distribution.

However, existing hierarchical \ac{DL} frameworks predominantly enforce a strictly \textit{top-down} information flow, where coarse-level predictions (\ac{FT}) act as hard masks or gating mechanisms for fine-level classifications (\ac{TS}). A critical flaw of this architectural design is the vulnerability to cascading errors: if the network misclassifies the parent category due to spectral ambiguities, the error inevitably propagates downward, trapping the model in the wrong taxonomic branch and making it impossible to recover the correct \ac{TS} \cite{silla2011survey}. The literature currently lacks frameworks capable of exploiting \textit{bidirectional} hierarchical learning, where sub-pixel species composition can simultaneously inform and solidify the broader category assignment. Our proposed method explicitly addresses this gap. By implementing a structurally-masked attention mechanism, the network supports bidirectional interaction between coarse and fine-grained representations, while the explicit ecological constraints applied at inference prevent invalid \ac{FT}-\ac{TS} combinations. The attention mechanism enables bottom-up feature aggregation, where fine-grained compositional features dynamically refine the macroscopic \ac{FT} prediction, while the deterministic hierarchy provides structural consistency relative to the encoded ecological relationships.

%% file: chapters/methodology.tex
In this work, we address the aforementioned challenges by proposing a novel \ac{DL} framework for sub-pixel-level classification of \ac{TS} that combines CNN and attention paradigms. The proposed method explicitly models the ecological hierarchical relationship between \ac{FT} and \ac{TS} (formalized in Section \ref{sec:hier-formulation}) and jointly leverages hyperspectral data, panchromatic imagery, and \ac{DEM}. Our approach is based on a multi-branch \ac{CNN} feature extractor (detailed in Section \ref{sec:feature_extractor}), where each branch processes a distinct type of input data. The network is designed to perform classification at two hierarchical levels.
The first level corresponds to \ac{FT} classification, while the second level refines the prediction to identify \ac{TS} using a structurally-masked Transformer architecture (presented in Section \ref{sec:ft_ts_classifier}). For each pixel, up to three co-occurring \ac{TS} can be detected, with associated ordinal rankings that reflect their relative ecological dominance within the forest stand.

The design of this two-level subdivision naturally reflects the taxonomic relationship between \ac{FT} and \ac{TS}. However, it is also driven by practical constraints in ground-truth data acquisition. Annotating \acp{FT} is relatively cost-effective because it primarily relies on visual photointerpretation and limited field campaigns. This approach facilitates the creation of large-scale reference datasets. In contrast, acquiring reliable \ac{TS} labels is highly labor-intensive. It requires detailed in situ surveys for precise species identification at the pixel level, which inevitably results in a severe scarcity of fine-grained annotated data.

To overcome the aforementioned data scarcity and effectively optimize the network, we introduce a decoupled, two-step training strategy (described in Section \ref{sec:training_strategy}). Initially, the multi-branch feature extractor and a preliminary \ac{FT} classification head are optimized using a large dataset containing only coarse \ac{FT} labels. Subsequently, the feature extractor weights are frozen, and a smaller, fully annotated dataset (composed of both \ac{FT} and \ac{TS} labels) is utilized to train a lightweight joint \ac{FT}-\ac{TS} Transformer, whose small number of parameters structurally prevents overfitting when learning complex ecological rules from limited \ac{TS} data. In this second phase, the structurally-masked attention mechanism plays a dual role: it leverages the macroscopic \ac{FT} context to guide the fine-grained \ac{TS} predictions \textit{top-down}, while simultaneously aggregating sub-pixel \ac{TS} assemblage features to refine and solidify the final \ac{FT} classification \textit{bottom-up}.

\subsection{Feature Extractor Architecture}
\label{sec:feature_extractor}
\begin{figure*}
    \centering
    \includegraphics[width=\linewidth]{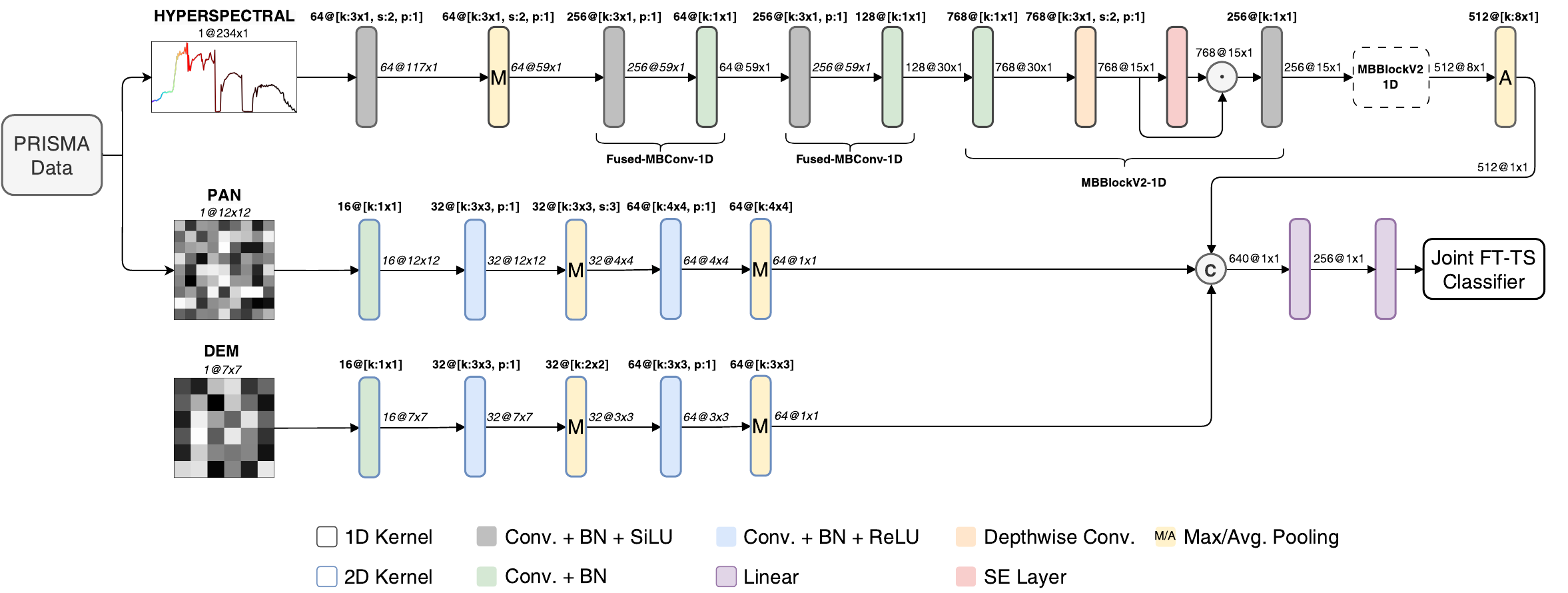}
    \caption{Overview of the proposed architecture for joint feature extraction and hierarchical classification. Structural specifications are reported as $C @ (H \times W)$ or $C @ L\times1$, indicating the number of channels and the spatial or spectral resolution of the feature maps, respectively. Layer-specific parameters are denoted by the triplet ($k, s, p$), representing kernel size, stride, and padding. Operational symbols $\cdot$ and $C$ denote scalar multiplication and feature concatenation, respectively. Acronyms: SE (Squeeze-and-Excitation), BN (Batch Normalization), M (Max Pooling), and A (Average Pooling).}
    \label{fig:architecture}
\end{figure*}

The feature extractor is structured as a multi-branch \ac{CNN}. A schematic overview of the proposed architecture, including the detailed configuration of its layers, is illustrated in Fig. \ref{fig:architecture}. The first branch is implemented as a one-dimensional \ac{CNN} designed to extract features from the spectral signature of each sample. Its structure is derived from the EfficientNetV2 architecture \cite{tan2021efficientnetv2} and reformulated for spectral data. The network receives a one-dimensional input and processes it through a stem convolution with 64 output channels, followed by batch normalization, Sigmoid Linear Unit (SiLU) activation, and a max-pooling operation that reduces the spectral resolution, leveraging the redundancy among contiguous bands to decrease the model complexity. The network then comprises four convolutional stages based on mobile inverted bottleneck convolution (MBConv) blocks. The first stage employs Fused-MBConv-1D blocks with an expansion ratio of 4 and 64 output channels, preserving resolution while enriching feature representations. The second stage increases the dimensionality to 128 channels and applies a stride of 2 for downsampling. The third and fourth stages are composed of MBConvV2-1D blocks with an expansion ratio of 6, incorporating depthwise separable convolutions and Squeeze-and-Excitation (SE) mechanisms that adaptively recalibrate channel responses. These deeper stages expand the channel depth to 256 and 512, respectively, capturing progressively higher-level spectral patterns. Residual connections are preserved whenever input and output dimensions are compatible, and a stochastic depth strategy randomly drops residual paths during training to improve regularization and robustness. An adaptive average pooling layer aggregates the final feature maps into a compact 512-dimensional feature representation. The choice of this type of branch is motivated by the intrinsic efficiency of the EfficientNetV2 architecture, which achieves high accuracy with a relatively small number of parameters. Given the limited training data, this compact design is specifically tailored to alleviate the risk of overfitting while supporting effective learning, rendering self-supervised pretraining a non-essential step.

The second and third branches are implemented as compact 2D \acp{CNN} designed to extract complementary spatial information from the panchromatic and \ac{DEM} representations of the hyperspectral samples, respectively. Each network receives a single-channel input that is first processed by a linear projection across channels followed by batch normalization to expand the number of channels to 16. Subsequently, two sequential convolutional blocks extract spatial features. Each block consists of a convolution, batch normalization, Rectified Linear Unit (ReLU) activation, and a max-pooling operation. In the first block, the number of channels increases from 16 to 32, while in the second block it expands from 32 to 64. Both branches produce compact feature representations of size 64. The shallow architecture of these two networks reflects their auxiliary purpose. While the primary discriminative power relies on the main hyperspectral branch, the \ac{DEM} models local topographic conditions, and the panchromatic imagery captures localized structural cues such as canopy roughness.

Once features are extracted from all three branches, they are concatenated to form a unified representation. To promote cross-modal interaction and ensure the subsequent Transformer module remains strictly lightweight, a linear layer reduces this concatenated feature vector from 640 to 256 dimensions. This fused representation is then fed into the joint FT-TS classifier detailed in Section \ref{sec:ft_ts_classifier}.

\subsection{Definition of the Hierarchical Classification Model}
\label{sec:hier-formulation}

The proposed hierarchical formulation, derived from the INFC (Inventario Nazionale delle Foreste e dei serbatoi forestali di Carbonio) categorization \cite{gasparini2022definitions}, is organized into two distinct levels: a coarse level represented by the \ac{FT} and a finer level represented by the \ac{TS}. The formulation complies with the following ecological constraints:

\begin{enumerate}[i)]
\item A \ac{TS} may occur in multiple \acp{FT}.
\item A \ac{TS} can be dominant (defined as the species exhibiting the highest prevalence within a pixel) in only one \ac{FT}, while it may appear as a second- and third-ranked component in several other \acp{FT}.
\item An \ac{FT} may encompass multiple dominant \acp{TS}.
\end{enumerate}

\begin{figure}
    \centering
    \includegraphics[width=\linewidth]{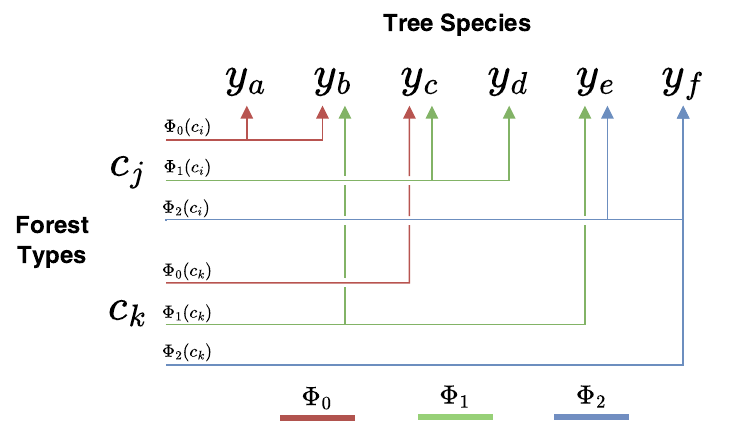}
    \caption{Illustration of the hierarchical formulation and ecological constraints derived from the INFC categorization. For a specific \ac{FT} $c_i$, the mapping function $\Phi$ identifies primary ($\Phi_0$) and secondary ($\Phi_1, \Phi_2$) \ac{TS}.}
    \label{fig:hier}
\end{figure}

A schematic representation of these hierarchical relationships is illustrated in Fig.~\ref{fig:hier}.

From these constraints, it follows that once a sample is assigned to a specific \ac{FT}, the set of admissible \acp{TS} is strictly bounded to those associated with that \ac{FT}. Conversely, constraint (ii) implies that if a \ac{TS} is identified as dominant, its corresponding \ac{FT} can be uniquely determined. While this bidirectional dependency conceptually allows inferring the coarse class from the fine-grained dominant species, our Transformer-based architecture absorbs this logic directly into its representation learning. Instead of applying hard \textit{bottom-up} corrections, the topology-constrained self-attention mechanism allows the \ac{FT} token to softly aggregate species-specific features from the \ac{TS} tokens during the forward pass. Meanwhile, the strict \textit{top-down} constraint (restricting admissible \acp{TS} based on the predicted \ac{FT}) is deterministically enforced during inference to guarantee structural validity.

Mathematically, this hierarchical structure is formalized by a mapping function $\Phi$, which associates each \ac{FT} class $c \in \mathcal{C}$ (with $|\mathcal{C}| = N_c$) with a valid set of \ac{TS} classes $y \in \mathcal{Y}$ (with $|\mathcal{Y}| = N_y$) across the three dominance levels ($l \in \{0,1,2\}$). Let $\Phi_l(c)$ $\subseteq \mathcal{Y}$ denote the subset of ecologically admissible \ac{TS} at dominance level $l$ for the \ac{FT} $c$.

According to our constraints, for a given \ac{FT} class $c_i$, the mapping at the dominant level yields a specific subset of primary species, e.g., $\Phi_0(c_i)=\{y_a, y_b\}$, while at the second and third levels it yields secondary components, e.g., $\Phi_1(c_i)=\{y_c, y_d\}$ and $\Phi_2(c_i)=\{y_e,y_f\}$, thus comprehensively formalizing the valid hierarchical structure of the pixel. Due to constraint (ii), the inverse mapping for the dominant level is unique for any given species; that is, if $y_a \in \Phi_0(c)$, then $c$ is uniquely identified ($\Phi_0^{-1}(y_a) = c_i$). Conversely, for lower dominance levels, a species may map back to multiple valid \acp{FT}, e.g., $\Phi_2^{-1}(y_f)=\{c_i, c_k\}$.

In practice, to natively integrate this structure into the neural architecture, this mapping is implemented as a binary transition tensor $\mathbf{H} \in \{0, 1\}^{N_c \times N_y \times 3}$, where $\mathbf{H}_{i, j, l} = 1$ if species $y_j$ is ecologically admissible at dominance level $l$ for \ac{FT} $c_i$ (i.e., $y_j \in \Phi_l(c_i)$), and $0$ otherwise.

\subsection{Joint FT-TS Classifier Architecture}
\label{sec:ft_ts_classifier}

The joint FT-TS classifier is designed to capture the complex, hierarchical dependencies between the \ac{FT} and the relative composition of the three co-occurring \ac{TS}. To achieve this, the module employs a structurally-masked Transformer architecture that explicitly integrates ecological transition rules.

Let $\mathbf{x} \in \mathbb{R}^{d}$ ($d=256$) represent the compressed feature vector resulting from the cross-modal fusion of the three branches (1D spectral, panchromatic, and DEM). To format this into a sequence suitable for the Transformer, $\mathbf{x}$ is duplicated and augmented with four condition-aware slot embeddings. Rather than using purely unconstrained learnable parameters, we explicitly condition these slots on the global ecological hierarchy. 

First, we compute a dense ecological embedding $\mathbf{e}_{H} \in \mathbb{R}^d$ by flattening the binary transition tensor $\mathbf{H}$ and projecting it through a learned linear layer:
\begin{equation}
    \mathbf{e}_{H} = \mathbf{W}_{H} \text{vec}(\mathbf{H}) + \mathbf{b}_{H},
\end{equation}
where $\text{vec}(\cdot)$ denotes the flattening operation, and $\mathbf{W}_{H} \in \mathbb{R}^{d \times (3 N_c N_y)}$ and $\mathbf{b}_{H} \in \mathbb{R}^d$ are the weights and bias of the projection layer, respectively. This ecological prior is then broadcast and added to the base learnable slot embeddings $\{\mathbf{s}_i\}_{i=1}^4 \subset \mathbb{R}^d$. The final hierarchy-aware input sequence $\mathbf{U} = [\mathbf{u}_1, \mathbf{u}_2, \mathbf{u}_3, \mathbf{u}_4]^\top \in \mathbb{R}^{4 \times d}$ fed to the Transformer is constructed such that each sequence element is defined as follows:
\begin{equation}
    \mathbf{u}_i = \mathbf{x} + \mathbf{s}_i + \mathbf{e}_{H} , \quad i = 1, \dots, 4.
\end{equation}
In this way, the attention mechanism is guided by prior ecological knowledge before dynamically interacting with the spectral features.

A preliminary classification head processes the first token (corresponding to \ac{FT}) to generate the initial \ac{FT} prediction $\hat{c}_{pre}$. As detailed in Section \ref{sec:training_strategy}, this preliminary output is specifically designed to provide the supervision signal necessary to train the multi-branch feature extractor during the first phase of our optimization strategy. The sequence $\mathbf{U}$ is then passed through a Transformer encoder layer. 

Notably, to enforce the desired flow of information, the self-attention mechanism is constrained by a hard topology mask $\mathbf{M} \in \mathbb{R}^{4 \times 4}$. The input sequence $\mathbf{U}$ is first projected into query ($\mathbf{Q}$), key ($\mathbf{K}$), and value ($\mathbf{V}$) matrices, all in $\mathbb{R}^{4 \times d}$, through learned linear transformations. In our formulation, the mask $\mathbf{M}$ is defined such that the \ac{FT} token (corresponding to the first index, $i=1$) can attend to all species tokens, thereby aggregating sub-pixel compositional information to refine its own representation. Conversely, the individual species tokens ($y_0, y_1, y_2$ at indices $i=2,3,4$) are restricted to attend only to themselves and the \ac{FT} token, preventing spurious cross-talk between species at this stage. The attention calculation is formalized as follows:
\begin{equation}
    \text{Attention}(\mathbf{Q}, \mathbf{K}, \mathbf{V}) = \text{Softmax}\left(\frac{\mathbf{Q}\mathbf{K}^T}{\sqrt{d}} + \mathbf{M}\right)\mathbf{V},
\end{equation}
where $\mathbf{M}_{i,j} = -\infty$ for masked connections and $0$ otherwise. The output of the Transformer is the sequence of refined tokens $\mathbf{Z} = [\mathbf{z}_c, \mathbf{z}_{y_0}, \mathbf{z}_{y_1}, \mathbf{z}_{y_2}]^\top \in \mathbb{R}^{4 \times d}$.

The final \ac{FT} prediction is derived directly from the refined token $\mathbf{z}_c$ via a dedicated linear classification head. Let $\mathbf{p}_c \in \mathbb{R}^{N_c}$ be the posterior probability distribution over the \ac{FT} classes, computed as:
\begin{equation}
\mathbf{p}_c = \text{Softmax}(\mathbf{W}_c \mathbf{z}_c + \mathbf{b}_c),
\end{equation}
where $\mathbf{W}_c \in \mathbb{R}^{N_c \times d}$ and $\mathbf{b}_c \in \mathbb{R}^{N_c}$ are the weights and bias of the classifier. The predicted class is then obtained as $\hat{c} = \text{argmax}(\mathbf{p}_c)$.

For the \ac{TS} classification, the module employs a sequential, cascaded prediction strategy. Although each prediction is intrinsically conditioned on the multimodal input $\mathbf{x}$, the \ac{FT} context $\mathbf{z}_c$, and the preceding steps of the cascade, for the sake of notational brevity we denote the posterior probability distributions simply as $\mathbf{p}_{y_l} \in \mathbb{R}^{N_y}$. First, the dominant species ($l=0$) is predicted directly from its refined token:

\begin{equation}
\mathbf{p}_{y_0} = \text{Softmax}(\mathbf{W}_{y_0} \mathbf{z}_{y_0} + \mathbf{b}_{y_0}).
\end{equation}

To ensure structural consistency during the forward pass, the predictions for the second- and third-ranked species are explicitly conditioned on the posterior probability distributions of the preceding dominant species. Specifically, the inputs to the $y_1$ and $y_2$ classification heads are constructed by concatenating their respective Transformer tokens with the softmax outputs of the prior predictions:
\begin{align}
    \mathbf{p}_{y_1} &= \text{Softmax}\big(\mathbf{W}_{y1} [\mathbf{z}_{y_1} \parallel \mathbf{p}_{y_0}] + \mathbf{b}_{y1}\big) \\
    \mathbf{p}_{y_2} &= \text{Softmax}\big(\mathbf{W}_{y2} [\mathbf{z}_{y_2} \parallel \mathbf{p}_{y_0} \parallel \mathbf{p}_{y_1}] + \mathbf{b}_{y2}\big)
\end{align}
where $\parallel$ denotes the concatenation operator.

\subsection{Training Strategy and Loss Formulation}
\label{sec:training_strategy}

The training process of the proposed architecture is divided into two distinct steps to effectively leverage heterogeneous data sources. 

In the first step, the feature extraction backbone and a preliminary \ac{FT} classification head are optimized using a larger dataset containing only coarse \ac{FT} annotations. The loss function for this step is defined as the categorical cross-entropy (CCE):
\begin{equation} 
    \mathcal{L}^c_{\text{CCE}} = - \frac{1}{N} \sum_{i=1}^{N} \mathbf{y}_{c,i} \cdot \log(\mathbf{p}_{\text{pre},i}),
\end{equation}
where $N$ is the number of training samples, $\mathbf{y}_{c,i}$ represents the one-hotencoded ground-truth \ac{FT} label, and $\mathbf{p}_{\text{pre},i}$ is the posterior probability distribution predicted by the preliminary classification head, from which the predicted class $\hat{c}_{\text{pre},i} = \text{argmax}(\mathbf{p}_{\text{pre},i})$ is derived. Once this step converges, the feature extractor is fully trained and provides robust baseline representations.

The second step focuses on training the joint FT-TS structured Transformer using a smaller dataset endowed with full hierarchical annotations. In this phase, the model is optimized via a multi-label loss function $\mathcal{L}_{\text{ml}}$ that simultaneously supervises the refined \ac{FT} output and the three cascaded \ac{TS} predictions. Specifically, the loss is formulated as a weighted combination of categorical cross-entropy terms:
\begin{equation} 
\begin{split}
    \mathcal{L}_{\text{ml}} = & - \frac{1}{N} \sum_{i=1}^{N} \Biggl( \alpha_c \mathbf{y}_{c, i} \cdot \log(\mathbf{p}_{c, i}) \\ 
    & + \sum_{l=0}^{2} \alpha_l \mathbf{y}_{y_l, i} \cdot \log(\mathbf{p}_{y_l,i}) \Biggr),
\end{split}
\end{equation}
where $\mathbf{y}_{y_l, i}$ are the ground-truth one-hot vectors for the $l$-th species level. The terms $\mathbf{p}_{c, i}$ and $\mathbf{p}_{y_l,i}$ denote the predicted posterior probability distributions generated by the four dedicated heads of the classifier. The hyperparameters $\alpha_c$ and $\alpha_l$ (for $l \in \{0, 1, 2\}$) serve as level-specific weights to balance the importance of the global \ac{FT} refinement against the fine-grained \ac{TS} composition. By backpropagating this joint loss through the topology-constrained attention mechanism, the network learns to mutually regularize the coarse and fine-grained predictions, ensuring that the final output is both ecologically consistent and spectrally accurate.

\begin{figure}
    \centering
    \includegraphics[width=\linewidth]{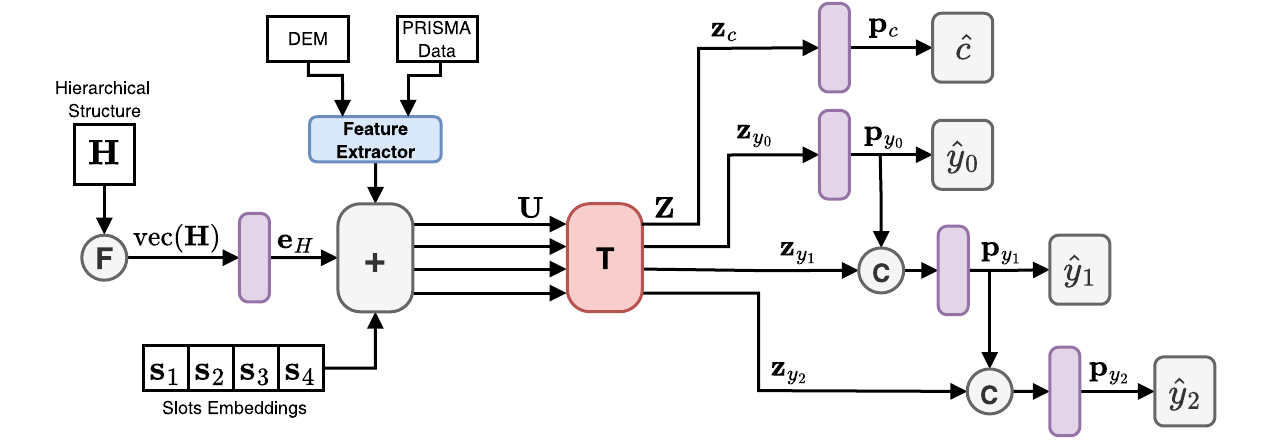}
    \caption{Overview of the architecture of the proposed Joint FT-TS Classifier. The pipeline highlights the integration of linear layers (violet), the Transformer block (T), and data reshaping operations, including flattening (F) and concatenation (C).}
    \label{fig:ft-ts}
\end{figure}

Finally, during inference, explicit deterministic constraints based on the hierarchy matrix $\mathbf{H}$ and a non-replacement rule are dynamically enforced. Any species $y_k$ deemed ecologically invalid for the predicted \ac{FT} $\hat{c}$, or already selected in a previous cascade step, is heavily penalized by setting its logit to $-\infty$. This post-processing guarantees a fully consistent and ecologically valid multi-label composition for the pixel. The overall workflow of the proposed algorithm is summarized in Fig. \ref{fig:ft-ts}.

%% file: chapters/dataset.tex
To evaluate the proposed model, we selected three study areas located in Italy characterized by high biogeographical diversity: the Alps and the Central and Southern Apennines. Unlike standard benchmarks, these datasets offer a rare alignment of regional hyperspectral images and detailed field-based composition data, enabling the validation of both \ac{FT} classification and sub-pixel \ac{TS} estimation.

\subsection{Study Areas}
The proposed methodology was evaluated over three ecologically distinct forested regions, selected to capture a wide range of biogeographical, altitudinal, and compositional variability. Together, these areas provide a challenging and representative testbed for detailed forest characterization in different conditions using spaceborne hyperspectral data.

The first study area is located in the central-eastern Italian Alps (Trentino-Alto Adige and Veneto regions), covering approximately 3,600 km$^2$. Elevation spans from lowland valleys (approximately 100 m above sea level (a.s.l.)) to high-alpine environments exceeding 2,500 m a.s.l. Forest composition follows pronounced altitudinal gradients, transitioning from deciduous broadleaf species at lower elevations to conifer-dominated stands at higher elevations, including subalpine and alpine forest types.

The second study area encompasses the Central Apennines (Lazio, Umbria, and Marche regions), also covering approximately 3,600 km$^2$. Elevation ranges from approximately 30 m to 2,200 m a.s.l. This region is characterized by a complex mosaic of Mediterranean, temperate, and montane forest ecosystems, reflecting both latitudinal and altitudinal ecological transitions.

The third study area is located in the Southern Apennines (Campania and Basilicata regions), covering approximately 2,700 km$^2$, with elevations ranging from 250 m to 1,900 m a.s.l. Forests in this region include Mediterranean and montane formations, with marked compositional variability across elevation belts.

\begin{figure}
    \centering
    \includegraphics[width=\linewidth]{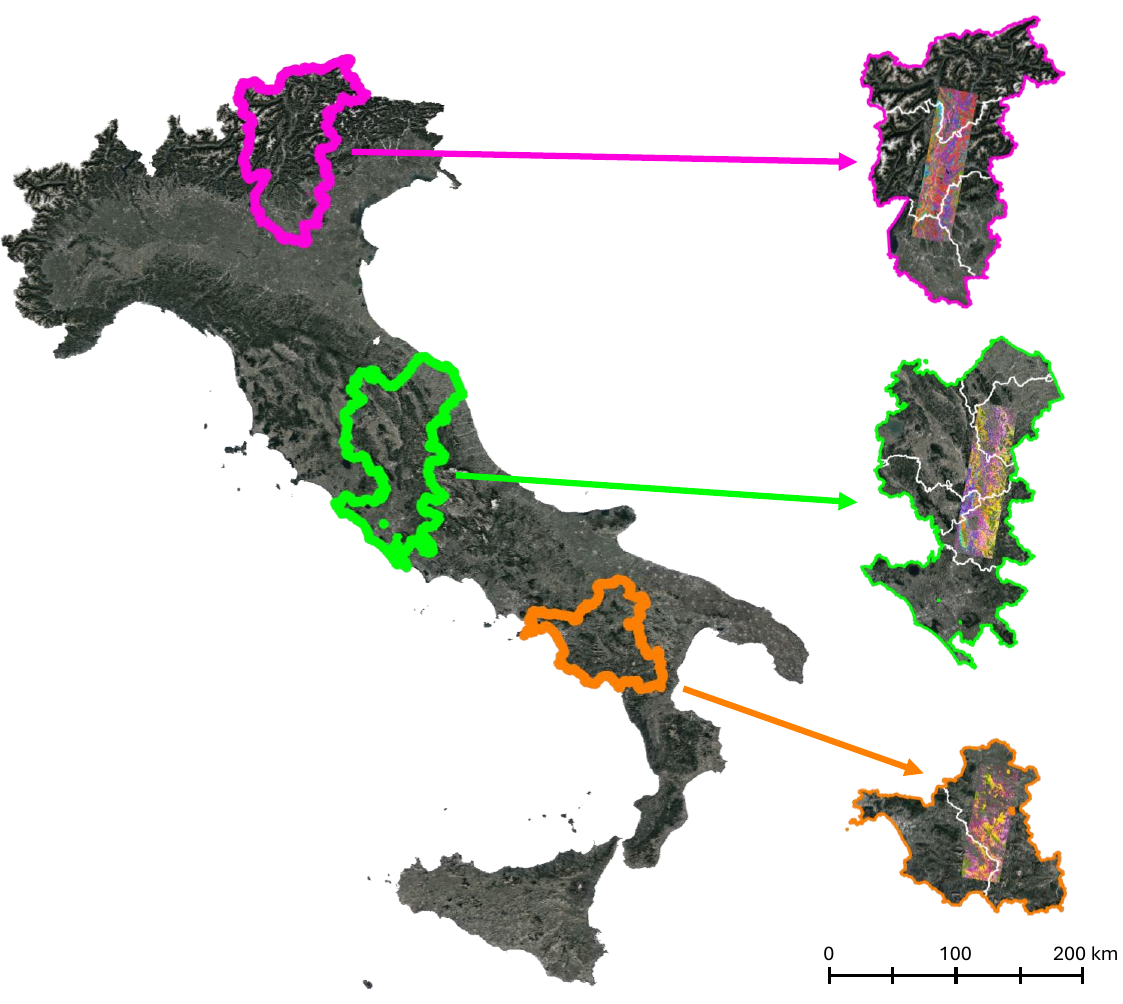}
    \caption{Geographical distribution of the three study areas across Italy, representing distinct forest ecosystems.}
    \label{fig:aoi}
\end{figure}

Fig. \ref{fig:aoi} shows the location of the three study areas. Table \ref{tab:n_samples} presents the different \acp{FT} present in each area and the corresponding number of ground-truth samples.

\subsection{Hyperspectral Data}
All study areas were covered by PRISMA hyperspectral imagery acquired during the summer of 2023, corresponding to peak vegetation development. A total of four images were used for both the central-eastern Alps and Central Apennines regions, and three images for the Southern Apennines region. All data were processed to Level 2D, i.e., geocoded bottom-of-atmosphere reflectance products.

PRISMA provides nominal spectral coverage from 400 to 1010 nm in the \ac{VNIR} and from 920 to 2505 nm in the \ac{SWIR}, with a spectral resolution of 10 nm, together with a panchromatic channel covering 400--700 nm \cite{candela2016prisma}. The hyperspectral and panchromatic products have native spatial resolutions of 30 m and 5 m, respectively. The hyperspectral sensors nominally record 239 channels (66 \ac{VNIR} and 173 \ac{SWIR}). Before concatenating the two hyperspectral cubes, we discarded the five uncalibrated, zero-valued bands (three in \ac{VNIR} and two in \ac{SWIR}), resulting in the 234 calibrated spectral bands used by the model.

Given the known geolocation uncertainty of PRISMA products \cite{de2024improving}, a geometric correction was applied to ensure spatial consistency between hyperspectral imagery and ground-truth data and to compensate for residual geolocation inaccuracies in PRISMA products. Image-to-reference co-registration was performed using the phase-based algorithm implemented in AROSICS (Automated and Robust Open-Source Image Co-registration Software) \cite{scheffler2017arosics}, aligning the PRISMA panchromatic band with high-resolution orthophotos provided by the AGEA (AGenzia per le Erogazioni in Agricoltura). This procedure ensured that ground-truth samples were correctly associated with their corresponding hyperspectral signatures. Radiometric normalization was performed independently for the \ac{VNIR} and \ac{SWIR} ranges using image-specific normalization coefficients provided in the PRISMA metadata.

\subsection{Ground-Truth and Auxiliary Data}
Ground truth information was derived from two complementary sources, providing both extensive spatial coverage and detailed compositional information.

The first ground-truth dataset was generated through expert photointerpretation \cite{rizzo2022land} and includes \ac{FT} labels only. It comprises 8,319 samples across 11 \acp{FT} in the central-eastern Alps region, 6,077 samples across 9 classes in the Central Apennines, and 2,869 samples across 5 classes in the Southern Apennines. This dataset supports robust training and evaluation of \ac{FT} classification performance. Class-wise sample distributions are reported in Table \ref{tab:n_samples}.

The second ground-truth dataset consists of detailed field measurements collected through in situ forest surveys, as comprehensively described in \cite{gasparini2025open}. As a post-processing step for this study, we refined the original field data by considering the \ac{FT} and up to three co-occurring species per plot. Their relative abundance was recorded as fractional cover in 10\% intervals, and species contributing less than 10\% were excluded from the analysis. It is important to highlight that not all \acp{FT} include a further division into \ac{TS}; therefore, the \ac{TS} analysis is restricted to the specific \ac{FT} categories where this information was available.

Auxiliary datasets included a \ac{DEM} and a forest mask. The \ac{DEM} was obtained from TINITALY at a 10-m spatial resolution \cite{tarquini2023tinitaly}. Forested areas were delineated using the Copernicus Forest Type 2021 Europe Map at a 10-m resolution \cite{copernicus2021}, ensuring that classification analyses were restricted to forest-covered pixels. This layer served as a spatial mask to generate the forest classification maps across all study areas.

\begin{table}[!t]
\centering
\caption{Number of ground-truth samples per forest category and study area.}
\label{tab:n_samples}
\renewcommand{\arraystretch}{1.1}
\footnotesize
\begin{tabular}{@{}l rrr @{}}
\toprule
\textbf{Forest Types} & \textbf{\shortstack{Central-East.\\Alps}} & \textbf{\shortstack{Central\\Apenn.}} & \textbf{\shortstack{South.\\Apenn.}} \\
\midrule
Larch and Swiss stone pine & 1,083 & -- & -- \\
Norway spruce & 1,124 & -- & -- \\
Fir & 1,088 & -- & -- \\
Scots / Mountain pine & 1,072 & -- & -- \\
Black pines & 409 & 671 & 94 \\
Mediterranean pines & -- & 465 & -- \\
Beech & 681 & 704 & 639 \\
Temperate oaks & 510 & 559 & -- \\
Mediterranean oaks & -- & 865 & 885 \\
Chestnut & 495 & 518 & 502 \\
Hornbeam / Hophornbeam & 768 & 579 & -- \\
Hygrophilous / Deciduous & 621 & -- & -- \\
Holm oak & -- & 1,066 & -- \\
No forest & 468 & 650 & 749 \\
\midrule
\textbf{TOTAL} & \textbf{8,319} & \textbf{6,077} & \textbf{2,869} \\
\bottomrule
\end{tabular}
\end{table}

%% file: chapters/setup.tex
To ensure a reliable evaluation despite the limited availability of \ac{TS} labels, we did not reserve a separate test set. Instead, we evaluated the models using 4-fold cross-validation independently for each study area, with task-specific fold construction as detailed below. Model performance is reported as the mean and standard deviation across the four validation folds.

\subsection{Proposed Data Splitting Strategy}

The photointerpreted samples used for \ac{FT} classification are partitioned using a spatially aware cross-validation strategy designed to prevent any possible spatial autocorrelation between training and validation data. First, for each \ac{FT}, the corresponding samples are grouped into spatial clusters using the $k$-means algorithm, with the number of clusters per class fixed to two.

\begin{figure}
    \centering
    \includegraphics[width=\linewidth]{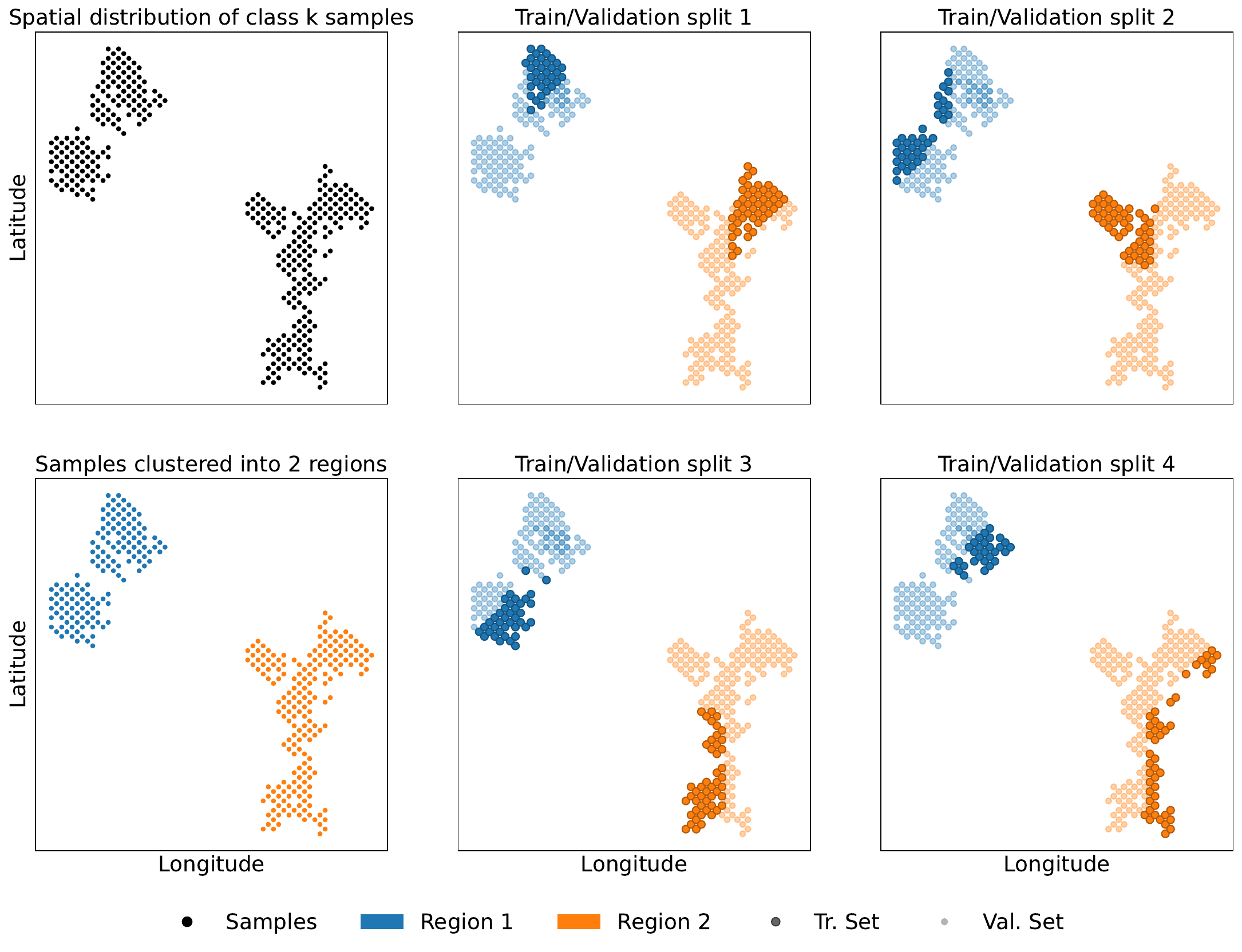}
    \caption{Illustration of the spatially aware cross-validation strategy for \ac{FT} classification. The spatial distribution of samples for a representative class $k$ (top-left) is first partitioned into two spatial clusters (bottom-left panel). An angular sector-based splitting algorithm is then applied to each cluster independently, generating four strictly disjoint train/validation splits (middle and right panels) to mitigate spatial autocorrelation.}
    \label{fig:ft_split}
\end{figure}

After clustering, the spatial centroid of each cluster is computed. The samples within each cluster are then sorted based on their angular position relative to this centroid. This angular sorting allows the samples to be seamlessly partitioned into four equal, spatially contiguous sectors. To implement a strict 4-fold cross-validation scheme, one distinct sector (representing 25\% of the samples of the cluster) is systematically allocated to the validation set in each fold, while the remaining three sectors (75\%) form the training set. Because this sector-based partitioning is applied independently to every cluster of each \ac{FT}, the strategy guarantees both reasonable spatial separation and consistent class representativeness across all four folds, as illustrated in Fig. \ref{fig:ft_split}.

For samples associated with \ac{TS} information, the complexity of ensuring that all possible multi-label combinations are adequately represented in both the training and validation sets, coupled with the strict requirements of a cross-validation framework, precludes the use of a spatially aware splitting strategy. Instead, a stratified 4-fold cross-validation scheme based on label clustering is employed. To ensure that varying species compositions are uniformly represented across all folds, the \ac{TS} labels are grouped into distinct strata using the $k$-means algorithm applied directly to the label space. Subsequently, within each formed cluster, the samples are randomly permuted and evenly partitioned into four disjoint subsets. Finally, in each fold, one distinct subset is systematically assigned as the validation set while the remaining three form the training set, guaranteeing that every sample is used for validation exactly once.

\subsection{Hyperparameter Optimization}
\label{sec:optimization}
Given the biogeographical differences and the distinct sets of forest categories present across the three study areas, model training and hyperparameter tuning procedures were conducted independently for each region.

For each specific area, experimental results are evaluated using the previously described 4-fold cross-validation strategy. Hyperparameter tuning is systematically performed using the Optuna framework, employing the Tree-structured Parzen Estimator (TPE) sampler \cite{bergstra2011algorithms}. During this process, the batch size is kept strictly fixed at $256$. The search space for both the learning rate and the weight decay is defined within the continuous range of $[10^{-5}, 10^{-2}]$, sampled on a logarithmic scale to efficiently explore varying orders of magnitude.

For the initial feature extraction and \ac{FT} classification module, the proposed multi-branch architecture was trained for 40 epochs. Furthermore, to provide a robust benchmark, we included two state-of-the-art architectures in our evaluation: SpectralFormer \cite{hong2022spectralformer} and Mamba \cite{pan2024hyperspectral}. To ensure a fair comparison, the configurations of these models were adjusted to achieve trainable parameter counts comparable to that of the proposed architecture. Due to their specific learning dynamics and to ensure proper convergence, the training process for these two comparison models required an extended duration of 150 epochs.

For the fine-grained \ac{TS} classification phase, the hyperparameter tuning strategy was strictly tailored to the specific architectural and optimization requirements of the evaluated models. Standard baseline models were trained for 150 epochs to ensure full convergence. 

In contrast, our proposed architecture demonstrated faster convergence and reached a stable plateau within 80 epochs per run. However, because our method introduces additional tunable hyperparameters, namely the weighting coefficients of the multi-label cross-entropy loss described in Section \ref{sec:training_strategy}, the dimensionality of its search space is significantly higher.

The optimization objective is tailored to the specific classification task. For the feature extraction module and the \ac{FT} classifier, the optimal hyperparameters are selected by maximizing the \ac{OA} on the validation folds. Conversely, the optimization of the \ac{TS} classifier is driven by the maximization of a composite metric, defined as the average of the per-level \ac{TS} accuracies and the \ac{FT} \ac{OA}.

To provide a robust estimate of the generalization capabilities of the model and stability, all final performance metrics are reported as the mean and standard deviation calculated across the four cross-validation folds.

\subsection{Evaluation Metrics}

The performance of the \ac{FT} classification task is evaluated using \ac{OA} and the F1-score, which jointly measure classification correctness and the balance between precision and recall.

The multi-label \ac{TS} classification task is evaluated using a comprehensive set of metrics:
\begin{itemize}
    \item \textbf{Per-level accuracy}, which measures the classification accuracy independently at each level of the hierarchical species labeling.
    \item \textbf{Composite accuracy}, computed as the average of the per-level accuracies and the \ac{FT} \ac{OA}, explicitly serving as the target metric maximized during the hyperparameter tuning of the \ac{FT}-\ac{TS} classifier.
    \item \textbf{Normalized Discounted Cumulative Gain (NDCG)}, which serves as a key ranking metric by assigning decaying relevance weights to the true classes (e.g., higher weight for the dominant species), heavily penalizing misranking errors at the top of the predicted distribution.
    \item \textbf{Mean Average Precision (mAP)}, which evaluates precision across different recall levels, rewarding models that place the true sub-pixel classes at the top of the predicted list, regardless of their strict hierarchical assignment.
    \item \textbf{Intersection over Union (IoU)}, which computes the set overlap between the true and predicted labels. While useful for assessing general co-occurrence, it is inherently limited in this context, as it yields a high score even if the hierarchical order (e.g., dominant vs. minor species) is completely inverted.
\end{itemize}

Together, these metrics provide a thorough evaluation of both classification accuracy and label ranking performance for the TS prediction task.

%% file: chapters/results_ft.tex
This section presents the experimental evaluation of the proposed hierarchical framework across the three distinct study areas (Central-Eastern Alps, Central Apennines, and Southern Apennines). To systematically assess the robustness and the specific contributions of the architecture, the analysis is organized into three progressive stages. First, Section \ref{sec:ft_classification} evaluates the coarse-grained \ac{FT} classification, validating the discriminative capability and the generalization power of the multi-branch feature extraction backbone and comparing it with two well-known approaches from the literature. Next, Section \ref{sec:comparison} constitutes the core of our analysis, benchmarking the joint \ac{FT}-\ac{TS} sub-pixel classification against representative multi-label and hierarchical paradigms to demonstrate the effectiveness of our structurally-masked Transformer. Finally, Sections \ref{sec:ablation_data} and  \ref{sec:ablation_components} detail the ablation studies conducted to isolate and quantify the impact of the multimodal data sources (hyperspectral, panchromatic, \ac{DEM}) and the individual architectural components, including the cascaded prediction, the self-attention mechanism, and the explicit ecological constraints.

\subsection{Baseline FT Classification and Feature Extraction Validation}
\label{sec:ft_classification}
The primary objective of this initial analysis is to validate the effectiveness of the feature extraction stage, establishing a clear baseline before evaluating the proposed methodology in Section \ref{sec:comparison}. Therefore, Tables \ref{tab:ft_results_alps}, \ref{tab:ft_results_central}, and \ref{tab:ft_results_south} report the \ac{OA} and F1-score for the standalone \ac{FT} classifier (denoted as ``Proposed''), which consists of the multi-branch feature extractor coupled with a simple linear classification layer. To comprehensively evaluate this baseline, we benchmark it against two established sequence modeling architectures: SpectralFormer \cite{hong2022spectralformer} and Mamba \cite{pan2024hyperspectral}. The proposed architecture is the configuration used to pretrain the feature extraction backbone of the comparison methods evaluated in Section \ref{sec:comparison}. Therefore, analyzing its performance provides a direct approach to assessing the underlying feature extraction capabilities independently of the \ac{TS} classification task.

For our proposed joint \ac{FT}-\ac{TS} framework, the feature extraction backbone is instead pretrained using the preliminary \ac{FT} classification head described in Section \ref{sec:ft_ts_classifier}. The \ac{FT} classification accuracy obtained by this preliminary head is highly consistent with the proposed model, with marginal fluctuations from -1.52\% to +0.57\%. This minimal variation proves that the feature extraction stage is trained with comparable effectiveness in both scenarios. Consequently, any performance gain observed in the \ac{TS} classification task (discussed in Section \ref{sec:comparison}) can be safely attributed strictly to the proposed methodology, rather than to a better-trained feature extractor.

\definecolor{Gallery}{rgb}{0.950,0.950,0.950}
\begin{table}[t]
\centering
\caption{Comparison of FT classification accuracy in the Central-Eastern Alps study area.}
\label{tab:ft_results_alps}
\begin{tblr}{
  width = \linewidth,
  colspec = {Q[c,m,1.8cm] X[1.5,l] X[c] X[c]},
  cells = {m, font=\scriptsize},
  row{1} = {font=\bfseries\scriptsize, c},
  hline{1,Z} = {0.8pt},
  hline{2} = {0.4pt},
  cell{2}{3} = {Gallery,c},
  cell{2}{4} = {Gallery,c},
}
No. of Classes & Model & OA (\%) & F1-score (\%) \\
11 & Proposed           & \textbf{83.42 $\pm$ 4.63} & \textbf{84.15 $\pm$ 4.55} \\
11 & SpectralFormer \cite{hong2022spectralformer} & \uline{$80.78 \pm 3.18$} & \uline{$80.02 \pm 3.33$} \\
11 & Mamba \cite{pan2024hyperspectral} & $76.38 \pm 5.33$ & $75.39 \pm 5.20$  \\
\end{tblr}
\end{table}

\begin{table}[t]
\centering
\caption{Comparison of FT classification accuracy in the Central Apennines study area.}
\label{tab:ft_results_central}
\begin{tblr}{
  width = \linewidth,
  colspec = {Q[c,m,1.8cm] X[1.5,l] X[c] X[c]},
  cells = {m, font=\scriptsize},
  row{1} = {font=\bfseries\scriptsize, c},
  hline{1,Z} = {0.8pt},
  hline{2} = {0.4pt},
  cell{2}{3} = {Gallery,c},
  cell{2}{4} = {Gallery,c},
}
No. of Classes & Model & OA (\%) & F1-score (\%) \\
9  & Proposed           & \textbf{88.10 $\pm$ 4.66} & \textbf{88.57 $\pm$ 4.37} \\
9  & SpectralFormer \cite{hong2022spectralformer} & \uline{$84.03 \pm 4.47$} & \uline{$83.76 \pm 4.45$} \\
9  & Mamba \cite{pan2024hyperspectral}          & $82.81 \pm 2.57$ & $82.69 \pm 2.17$ \\
\end{tblr}
\end{table}

\begin{table}[t]
\centering
\caption{Comparison of FT classification accuracy in the Southern Apennines study area.}
\label{tab:ft_results_south}
\begin{tblr}{
  width = \linewidth,
  colspec = {Q[c,m,1.8cm] X[1.5,l] X[c] X[c]},
  cells = {m, font=\scriptsize},
  row{1} = {font=\bfseries\scriptsize, c},
  hline{1,Z} = {0.8pt},
  hline{2} = {0.4pt},
  cell{2}{3} = {Gallery,c},
  cell{2}{4} = {Gallery,c},
}
No. of Classes & Model & OA (\%) & F1-score (\%) \\
5  & Proposed           & \textbf{98.16 $\pm$ 1.15} & \textbf{97.85 $\pm$ 1.47} \\
5  & SpectralFormer \cite{hong2022spectralformer} & $96.46 \pm 2.16$ & $96.14 \pm 2.49$ \\
5  & Mamba \cite{pan2024hyperspectral}          & \uline{$96.89 \pm 1.68$} & \uline{$96.56 \pm 2.26$} \\
\end{tblr}
\end{table}

Overall, the \ac{FT} classification results reveal different levels of classification difficulty across the three study areas, broadly reflecting their taxonomic complexity and spectral separability. The Central-Eastern Alps area represents the most complex scenario (achieving a maximum of 83.42\% \ac{OA}) primarily due to the highest number of target classes (11 distinct categories, as detailed in Table \ref{tab:n_samples}). The necessity to distinguish among several closely related and spectrally similar \ac{FT}s inherently amplifies the spectral confusion. The Central Apennines show a more moderate complexity with 9 classes, yielding an intermediate performance (up to 88.10\% \ac{OA}). Finally, the Southern Apennines dataset presents the most favorable conditions for spectral separability, coupled with a simplified taxonomic scheme (only 5 categories), allowing the best model to peak at 98.16\% \ac{OA}.

Across all three taxonomic scenarios, the proposed model consistently outperforms both SpectralFormer and Mamba architectures. In the highly challenging Central-Eastern Alps dataset, our multi-branch baseline surpasses SpectralFormer and Mamba by margins of 2.64\% and 7.04\% in \ac{OA}, respectively. A similar hierarchy is maintained in the Central Apennines, where the proposed model preserves a solid margin over the attention-based and state-space competitors. In the more favorable Southern Apennines dataset, all methods converge toward high classification accuracy (exceeding 96\% \ac{OA}); nevertheless, the proposed model remains superior, while Mamba proves slightly more effective than SpectralFormer. This consistent superiority highlights that the proposed multi-branch CNN structure is more effective than the sequential models used for comparison at extracting discriminative spatial-spectral features for fine-grained vegetation classification.

Despite this expected variability across regions and model capacities, it is crucial to note that for each study area, the F1-scores closely mirror their respective \ac{OA} values for all compared methods (e.g., 84.15\% F1 vs. 83.42\% \ac{OA} for the proposed model in the Central-Eastern Alps area). This tight alignment demonstrates that the tested frameworks effectively handle the class imbalances present in the datasets, avoiding majority-class bias and ensuring robust, reliable predictions across all individual \ac{FT}s.

\begin{figure*}
    \centering
    \includegraphics[width=\linewidth]{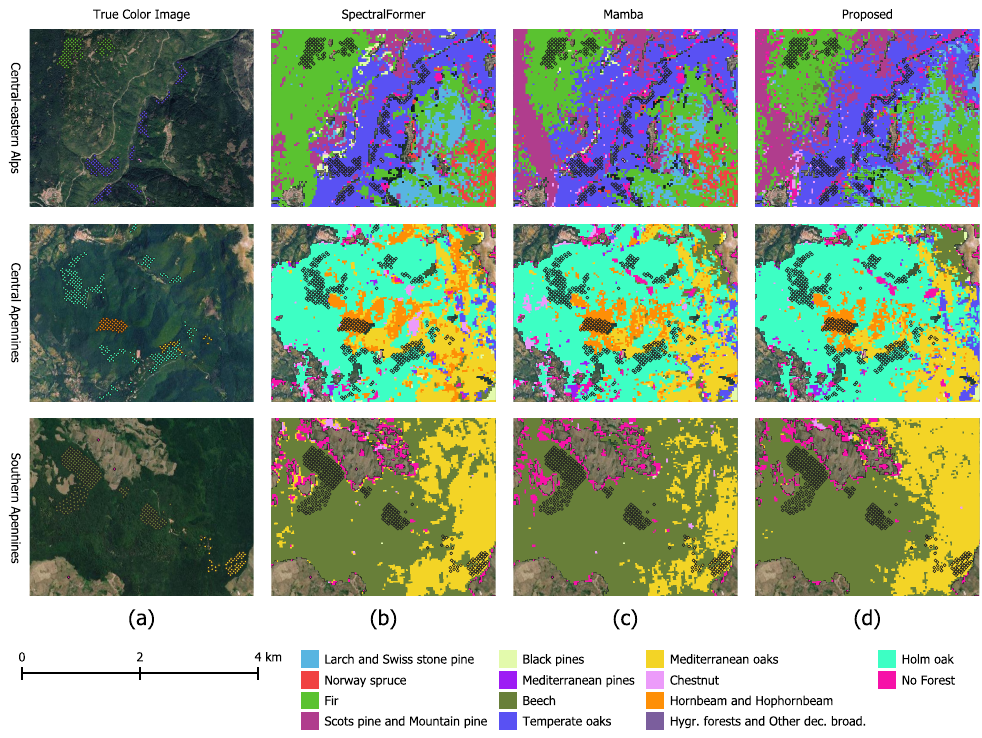}
    \caption{Representative subset of the generated FT classification maps for the three study areas: Central-Eastern Alps, Central Apennines, and Southern Apennines (ordered from top to bottom). (a) Reference true-color image. (b)–(d) Classification results for the proposed, SpectralFormer, and Mamba architectures, respectively. Overlaid markers represent ground-truth samples.}
    \label{fig:cl_maps}
\end{figure*}

Finally, a subset of the generated classification maps for the three scenarios is depicted in Fig. \ref{fig:cl_maps}. Across all study areas, the three classification models exhibit high spatial consistency with the ground-truth observations. In the central-eastern Alps, the generated maps are largely comparable; the primary discrepancy lies in the detection of Scots pine and Mountain pine, which are identified by both the proposed model and Mamba in areas where SpectralFormer predicts Fir. Additionally, the proposed model demonstrates a superior delineation of the ``No Forest'' class, a qualitative improvement that is evident across all three regions. In the Central Apennines, minor variations emerge in the northeastern sector of the maps. Here, the proposed model and Mamba again show similar spatial distributions, whereas SpectralFormer predicts a higher prevalence of Hornbeam and Hophornbeam rather than Beech, a confusion given the high similarity of their spectral signatures. Finally, in the Southern Apennines scenario, the classification maps are highly consistent, exhibiting only marginal differences in the relative proportions of Mediterranean oaks and Beech.

%% file: chapters/results_ts.tex
\subsection{Sub-Pixel Tree Species Classification}
\label{sec:comparison}
To the best of our knowledge, there are no papers in the literature that simultaneously address the intersection of hierarchical constraints and ordered multi-label sub-pixel classification. Consequently, a direct comparison with an identical end-to-end framework is infeasible. To rigorously evaluate our methodology, we benchmark it against a selection of representative baselines and state-of-the-art paradigms. These methods tackle individual aspects of the problem, such as sub-pixel ranking or hierarchical conditioning. However, they lack the joint ecological and structural formulation proposed in our approach.

Wherever compatible with their original formulations, the comparison methods use the same multi-branch \ac{CNN} feature extraction backbone as the proposed approach (fusing hyperspectral, panchromatic, and \ac{DEM} inputs), unless explicitly stated otherwise. For the complete ordered multi-label baselines, this common backbone focuses the comparison on the \ac{TS} prediction paradigm, loss function, and treatment of label dependencies. Methods whose original output is a single species label are instead treated separately as partial baselines.

As a foundational baseline, a standard \textit{Binary Relevance} approach optimized via binary cross-entropy (BCE) loss is evaluated \cite{zhang2018binary}. In this setting, each label is predicted independently using a sigmoid-activated output. During inference, labels are ranked according to their predicted confidence scores to extract the top-3 classes. While computationally straightforward and widely adopted, BCE completely ignores the relative dominance ordering of the predicted labels and fails to model the intrinsic ecological dependencies between the broad \ac{FT} macro-classes and the fine-grained \ac{TS} mixtures. As a result, neither sub-pixel ranking quality nor taxonomic consistency is explicitly enforced.

To address the limitations of independent label optimization for mixed-species canopies, two ranking-based loss functions are incorporated into the comparison. The first is \textit{ListNet} \cite{cao2007learning}, a listwise ranking method that models a probability distribution over label permutations, minimizing the divergence between the predicted and ground-truth ranking distributions. By operating at the list level, ListNet explicitly encourages the correct relative ordering among co-occurring species. The second approach employs the \textit{Approximate Normalized Discounted Cumulative Gain (ApproxNDCG)} loss \cite{qin2010general}, which provides a differentiable approximation of the NDCG metric. This loss emphasizes the correct ranking of the dominant species by assigning exponentially higher penalties to misranking errors occurring at the top of the predicted distribution. However, while listwise losses are theoretically designed to optimize the overall sub-pixel ranking, they remain entirely agnostic to the hierarchical relationships and the ecological transition rules between \ac{FT} and \ac{TS}.

An alternative paradigm for handling structured label dependencies is the \textit{Label Powerset} approach \cite{tsoumakas2007multi}. This method reformulates the multi-label problem as a standard multi-class classification task, where each class corresponds to a unique combination of labels across the different levels of species dominance. Rather than predicting multiple interacting labels and explicitly learning their relative rank, the model predicts a single compound class representing a specific ordered mixture of co-occurring species. Although this strategy rigidly captures species co-occurrence patterns within a pixel, it treats these complex mixtures as flat, independent categories, thereby failing to natively exploit the ordinal relationships among the species. Furthermore, it suffers from significant scalability issues, as the combinatorial space of possible sub-pixel compositions expands exponentially in highly diverse forest landscapes.

Finally, the proposed architecture is compared against methods designed to explicitly exploit taxonomic hierarchies. We evaluate two distinct approaches: the \textit{Switcher-HNet} and a multi-head \textit{Hierarchical} classifier. The \textit{Switcher-HNet} performs a sequential coarse-to-fine classification: it first discriminates between \ac{LT} categories using an independent binary classifier, and subsequently predicts the specific \ac{TS} conditioned exclusively on this first-level hard decision \cite{liu2026switcher}. To implement this hard-routing stage, we trained an independent binary \ac{LT} classifier distinguishing broadleaves from conifers. This auxiliary classifier serves only as the coarse routing component of Switcher-HNet and is therefore not directly comparable with the multi-class \ac{FT} results reported in Tables \ref{tab:ft_results_alps}--\ref{tab:ft_results_south}. It achieved OA and F1-score values of $98.23 \pm 1.02$\% and $98.15 \pm 1.06$\% in the Central-Eastern Alps and $97.67 \pm 1.32$\% and $96.09 \pm 2.23$\% in the Central Apennines. Conversely, the Hierarchical baseline \cite{sen2022hierarchical} shares the same low-level multimodal feature extraction backbone as our proposed method but employs a multi-head architecture. Specifically, separate linear classifiers are dedicated to different hierarchical levels (i.e., coarse \ac{FT} and fine-grained \ac{TS}). While this allows the network to capture both general and fine-grained distinguishing features using dedicated learning capacities, the independent heads do not enforce an ordered multi-label ranking. It is worth noting that both methods are natively formulated to predict a single species label and therefore do not explicitly model the ordered composition of co-occurring species. To obtain a common ranked output for evaluation, we retain the three species with the highest posterior probabilities and interpret them as $(\hat{y}_0,\hat{y}_1,\hat{y}_2)$, while preserving fine-grained species supervision exclusively on the dominant-species label $y_0$. These variants are consequently regarded as partial hierarchical baselines, since subordinate species are obtained through post-hoc posterior ranking rather than explicitly learned from $y_1$ and $y_2$. The resulting ranked predictions allow the computation of per-level accuracies and ranking-based metrics, providing an additional characterization of their behavior on mixed-species pixels. However, these results must be interpreted as diagnostic evaluations of single-label classifiers rather than as a fully task-matched comparison with methods explicitly trained for ordered multi-label classification.

For compactness in the tables, \ac{FT} accuracy, composite accuracy, and per-level accuracy are abbreviated as FT Acc., Comp. Acc., and Acc., respectively.

\input{tables/ts_alpine}
\input{tables/ts_central}
\input{tables/ts_south}

The quantitative results of the multi-label sub-pixel classification are reported in Tables \ref{tab:results_n}, \ref{tab:results_c}, and \ref{tab:results_s} for the Central-Eastern Alps, Central Apennines, and Southern Apennines study areas, respectively. The proposed hierarchy-aware architecture is evaluated against the complete ordered multi-label baselines on the full task, while the hierarchical single-label methods are included as partial baselines. For the latter, dominant-species accuracy ($y_0$) represents the directly task-matched comparison, whereas the additional metrics characterize the ranked predictions obtained through post-hoc processing. Among the complete ordered multi-label baselines, the proposed method attains the highest composite accuracy, our primary target metric, in all three study areas.

A preliminary observation must be made regarding the coarse-level classification. The complete ordered multi-label baselines and the multi-head Hierarchical method rely on the static \ac{FT} predictions generated by the pretrained, multi-branch \ac{CNN} feature extractor, yielding identical baseline \ac{FT} accuracies across the respective study areas. In contrast, \textit{Switcher-HNet} employs a separate binary \ac{LT} classifier for its coarse routing stage; consequently, neither \ac{FT} accuracy nor composite accuracy is reported for this method, as its coarse-level output is not equivalent to the multi-class \ac{FT} prediction required by these metrics. In contrast, the proposed method leverages the structurally-masked Transformer to perform a \textit{bottom-up} refinement. By allowing the global \ac{FT} representation to integrate the fine-grained \ac{TS} compositional information during the forward pass, the proposed framework successfully improves the broad \ac{FT} classification, reaching 92.58\% and 91.46\% in the Central-Eastern Alps and Central Apennines areas, respectively. In the Southern Apennines area (Table \ref{tab:results_s}), where the baseline \ac{FT} accuracy is already near saturation (98.05\%), the proposed method records a slight accuracy decrease (97.57\%).
This marginal drop reflects an ecological trade-off. The Transformer's structural constraints act as a regularizer: they slightly penalize coarse macro-class accuracy to improve the representation of highly intermixed subordinate species. This regularization translates into a tangible improvement at the deepest and most challenging compositional level ($y_2$).

When analyzing the fine-grained species classification, the limitations of standard multi-label paradigms become evident. The \textit{Binary Relevance} baseline shows a noticeable performance drop as the compositional depth increases toward the secondary components ($y_1$, $y_2$). Interestingly, ranking-based losses, such as \textit{Approx. NDCG} and \textit{ListNet}, fail to mitigate this issue at deeper levels. While they successfully optimize the dominant species ($y_0$) and overall continuous metrics (e.g., mAP) by penalizing top-tier misranking, they often suffer a more severe degradation on $y_1$ and $y_2$ compared to \textit{Binary Relevance} (e.g., \textit{ListNet} dropping to 34.87\% on $y_2$ in the Alps). Being fundamentally unaware of the taxonomic hierarchy, these losses tend to sacrifice highly intermixed subordinate species to secure the top-ranked prediction. Consequently, the complete proposed classifier, which combines cascaded prediction, topology-constrained attention, and ecological conditioning, achieves the strongest overall per-level performance on the challenging $y_1$ and $y_2$ classes; the contributions of these components are examined separately in Section \ref{sec:ablation_components}.

The evaluation of the \textit{Label Powerset} baseline provides a critical insight into the nature of sub-pixel forest heterogeneity. Across all regions, and most notably in the Southern Apennines, the \textit{Label Powerset} achieves the highest IoU scores (up to 79.61\%), yet it struggles to predict the specific subordinate classes accurately (e.g., dropping to 17.50\% on $y_2$ in the Southern Apennines). This apparent paradox highlights the inherent limitations of reformulating ordered multi-label tasks into flat multi-class problems: while the model effectively memorizes the most frequent exact combinations (boosting global overlap metrics), it fundamentally lacks the structural awareness required to resolve deep, diverse compositional mixtures.

Regarding the hierarchy-exploiting baselines, the multi-head Hierarchical method achieves competitive $y_0$ accuracies (89.37\%, 92.23\%, and 98.02\% across the three study areas), whereas \textit{Switcher-HNet} reaches 80.13\% and 90.04\% in the Central-Eastern Alps and Central Apennines, respectively. However, both methods exhibit substantial degradation at subordinate levels. For instance, in the Central Apennines, the Hierarchical baseline achieves only 9.98\% and 22.42\% on $y_1$ and $y_2$, respectively, compared with 64.07\% and 50.12\% for the proposed method. Despite these differences, the hierarchical baselines retain high mAP values and, in some cases, comparable NDCG scores, suggesting that their predictions can capture relevant species information without accurately resolving the complete ordered composition. This interpretation is further supported by their consistently lower IoU scores, which reveal limitations in recovering the full set of co-occurring species. Overall, these results highlight the importance of explicitly modeling subordinate species, although the absence of multi-label supervision prevents a fully task-matched comparison.

Finally, it is worth noting that the \textit{Switcher-HNet} was explicitly excluded from the Southern Apennines evaluation (Table \ref{tab:results_s}). Due to the complete absence of coniferous \ac{TS} in this specific study area, its coarse \ac{LT} hard routing is rendered conceptually inapplicable.

Among the complete ordered multi-label methods evaluated, the proposed classifier attains the highest Composite Accuracy in all three study areas. Its Comp. Acc. values of 80.26\%, 74.59\%, and 79.81\% are the highest among the complete ordered multi-label baselines; the partial hierarchical baselines are excluded from this full-task comparison.

%% file: tables/ts_alpine.tex
\definecolor{Gallery}{rgb}{0.950,0.950,0.950}
\begin{table*}[t]
\centering
\caption{Performance comparison on the Central-Eastern Alps study area. All values are percentages (\%), except for NDCG. ``--'' denotes unavailable metrics.}
\label{tab:results_n}
\begin{tblr}{
  width = \linewidth,
  colspec = {X[1.5,l] *{8}{X[c]}},
  cells = {m, font=\scriptsize},
  row{1} = {font=\bfseries\scriptsize, c},
  hline{1,Z} = {0.8pt},
  hline{2} = {0.4pt},
  cell{8}{2} = {Gallery,c},
  cell{8}{3} = {Gallery,c},
  cell{8}{4} = {Gallery,c},
  cell{8}{5} = {Gallery,c},
  cell{8}{6} = {Gallery,c},
  cell{3}{7} = {Gallery,c},
  cell{4}{8} = {Gallery,c},
  cell{5}{9} = {Gallery,c}
}
Method & FT Acc. & Comp. Acc. & Acc. $y_0$ & Acc. $y_1$ & Acc. $y_2$ & NDCG & mAP & IoU \\
Binary Relevance & \uline{91.02 $\pm$ 2.75} & 70.70 $\pm$ 4.19 & 83.01 $\pm$ 4.34 & 59.65 $\pm$ 2.33 & \uline{49.12 $\pm$ 13.6} & 0.614 $\pm$ 0.012 & \uline{97.43 $\pm$ 1.67} & 57.83 $\pm$ 1.12 \\
Approx. NDCG & \uline{91.02 $\pm$ 2.75} & \uline{72.17 $\pm$ 3.80} & 89.21 $\pm$ 0.96 & \uline{62.86 $\pm$ 4.94} & 45.58 $\pm$ 9.14 & \textbf{0.640 $\pm$ 0.008} & 96.93 $\pm$ 0.89 & 55.36 $\pm$ 1.48 \\
ListNet & \uline{91.02 $\pm$ 2.75} & 68.51 $\pm$ 1.72 & \uline{90.09 $\pm$ 1.60} & 58.08 $\pm$ 5.80 & 34.87 $\pm$ 3.84 & \uline{0.639 $\pm$ 0.010} & \textbf{97.79 $\pm$ 1.89} & 54.01 $\pm$ 0.91 \\
Label Powerset & \uline{91.02 $\pm$ 2.75} & 70.82 $\pm$ 1.97 & 89.94 $\pm$ 3.01 & 63.01 $\pm$ 2.71 & 39.34 $\pm$ 2.23 & 0.611 $\pm$ 0.010 & 96.56 $\pm$ 1.68 & \textbf{77.78 $\pm$ 2.68} \\
Hierarchical & \uline{91.02 $\pm$ 2.75} & 64.78 $\pm$ 4.17 & 89.37 $\pm$ 2.70 & 35.20 $\pm$ 5.74 & 43.53 $\pm$ 6.91 & 0.599 $\pm$ 0.013 & 94.90 $\pm$ 1.47 & 47.53 $\pm$ 1.67 \\
Switcher-HNet & -- & -- & 80.13 $\pm$ 1.74 & 35.76 $\pm$ 1.78 & 21.71 $\pm$ 6.85 & 0.576 $\pm$ 0.008 & 90.15 $\pm$ 2.93 &  43.13 $\pm$ 2.02  \\
\textbf{Proposed} & \textbf{92.58 $\pm$ 1.21} & \textbf{80.26 $\pm$ 2.67} & \textbf{91.01 $\pm$ 1.34} & \textbf{74.94 $\pm$ 3.32} & \textbf{62.53 $\pm$ 11.13} & 0.638 $\pm$ 0.015 & 97.32 $\pm$ 1.13 & \uline{58.27 $\pm$ 0.80} \\
\end{tblr}
\end{table*}

%% file: tables/ts_central.tex
\definecolor{Gallery}{rgb}{0.950,0.950,0.950}
\begin{table*}[t]
\centering
\caption{Performance comparison on the Central Apennines study area. All values are percentages (\%), except for NDCG. ``--'' denotes unavailable metrics.}
\label{tab:results_c}
\begin{tblr}{
  width = \linewidth,
  colspec = {X[1.3,l] *{8}{X[c]}},
  cells = {m, font=\scriptsize},
  row{1} = {font=\bfseries\scriptsize, c},
  hline{1,Z} = {0.8pt},
  hline{2} = {0.4pt},
  cell{8}{2} = {Gallery,c}, 
  cell{8}{3} = {Gallery,c}, 
  cell{5}{4} = {Gallery,c}, 
  cell{8}{5} = {Gallery,c}, 
  cell{8}{6} = {Gallery,c},
  cell{8}{7} = {Gallery,c}, 
  cell{5}{8} = {Gallery,c}, 
  cell{5}{9} = {Gallery,c}  
}
Method & FT Acc. & Comp. Acc. & Acc. $y_0$ & Acc. $y_1$ & Acc. $y_2$ & NDCG & mAP & IoU \\
Binary Relevance & \uline{90.97 $\pm$ 0.87} & 72.72 $\pm$ 2.10 & 89.70 $\pm$ 3.48 & \uline{61.82 $\pm$ 2.99} & \uline{48.37 $\pm$ 5.79} & \uline{0.806 $\pm$ 0.021} & 98.47 $\pm$ 0.89 & \uline{65.80 $\pm$ 0.74} \\
Approx. NDCG & \uline{90.97 $\pm$ 0.87} & 71.44 $\pm$ 2.02 & 93.33 $\pm$ 2.57 & 57.87 $\pm$ 3.88 & 43.57 $\pm$ 5.67 & 0.793 $\pm$ 0.022 & 97.78 $\pm$ 0.83 & 59.71 $\pm$ 0.80 \\
ListNet & \uline{90.97 $\pm$ 0.87} & \uline{72.99 $\pm$ 2.90} & \uline{94.02 $\pm$ 3.14} & 59.85 $\pm$ 1.31 & 47.11 $\pm$ 6.48 & 0.804 $\pm$ 0.022 & \uline{98.12 $\pm$ 0.57} & 61.85 $\pm$ 1.12 \\
Label Powerset & \uline{90.97 $\pm$ 0.87} & 70.59 $\pm$ 2.23 & \textbf{95.23 $\pm$ 1.40} & 54.72 $\pm$ 3.20 & 41.43 $\pm$ 7.53 & 0.802 $\pm$ 0.016 & \textbf{98.53 $\pm$ 0.33} & \textbf{72.38 $\pm$ 1.89} \\
Hierarchical & \uline{90.97 $\pm$ 0.87} & 53.90 $\pm$ 3.70 & 92.23 $\pm$ 0.88 & 9.98 $\pm$ 5.65 & 22.42 $\pm$ 5.64 & 0.696 $\pm$ 0.024 & 96.08 $\pm$ 2.09 & 32.65 $\pm$ 2.97 \\
Switcher-HNet & -- & -- & 90.04 $\pm$ 1.32 & 39.54 $\pm$ 2.95 & 14.48 $\pm$ 6.09 & 0.756 $\pm$ 0.027 & 98.06 $\pm$ 0.29 & 45.83 $\pm$ 2.11 \\
\textbf{Proposed} & \textbf{91.46 $\pm$ 2.69} & \textbf{74.59 $\pm$ 0.98} & 92.72 $\pm$ 2.62 & \textbf{64.07 $\pm$ 1.39} & \textbf{50.12 $\pm$ 3.16} & \textbf{0.806 $\pm$ 0.023} & 97.76 $\pm$ 0.57 & 63.96 $\pm$ 1.68 \\
\end{tblr}
\end{table*}

%% file: tables/ts_south.tex
\definecolor{Gallery}{rgb}{0.950,0.950,0.950}
\begin{table*}[t]
\centering
\caption{Performance comparison on the Southern Apennines study area. All values are percentages (\%), except for NDCG. ``--'' denotes unavailable metrics.}
\label{tab:results_s}
\begin{tblr}{
  width = \linewidth,
  colspec = {X[1.3,l] *{8}{X[c]}},
  cells = {m, font=\scriptsize},
  row{1} = {font=\bfseries\scriptsize, c},
  hline{1,Z} = {0.8pt},
  hline{2} = {0.4pt},
  cell{2}{2} = {Gallery,c},
  cell{3}{2} = {Gallery,c},
  cell{4}{2} = {Gallery,c},
  cell{5}{2} = {Gallery,c},
  cell{6}{2} = {Gallery,c},
  cell{7}{3} = {Gallery,c}, 
  cell{4}{4} = {Gallery,c}, 
  cell{7}{5} = {Gallery,c}, 
  cell{7}{6} = {Gallery,c}, 
  cell{4}{7} = {Gallery,c}, 
  cell{5}{8} = {Gallery,c}, 
  cell{5}{9} = {Gallery,c}  
}
Method & FT Acc. & Comp. Acc. & Acc. $y_0$ & Acc. $y_1$ & Acc. $y_2$ & NDCG & mAP & IoU \\
Binary Relevance & \textbf{98.05 $\pm$ 0.20} & \uline{78.29 $\pm$ 3.19} & 97.19 $\pm$ 1.37 & \uline{72.22 $\pm$ 9.57} & 45.71 $\pm$ 4.40 & 0.469 $\pm$ 0.057 & 98.93 $\pm$ 0.69 & \uline{48.16 $\pm$ 1.41} \\
Approx. NDCG & \textbf{98.05 $\pm$ 0.20} & 73.66 $\pm$ 11.79 & 97.62 $\pm$ 0.63 & 51.82 $\pm$ 17.60 & \uline{47.14 $\pm$ 37.17} & \uline{0.469 $\pm$ 0.057} & 99.09 $\pm$ 0.87 & 45.18 $\pm$ 1.66 \\
ListNet & \textbf{98.05 $\pm$ 0.20} & 77.69 $\pm$ 2.52 & \textbf{98.40 $\pm$ 1.86} & 67.77 $\pm$ 5.38 & 46.55 $\pm$ 9.76 & \textbf{0.474 $\pm$ 0.063} & \uline{99.85 $\pm$ 0.15} & 47.98 $\pm$ 1.48 \\
Label Powerset & \textbf{98.05 $\pm$ 0.20} & 65.70 $\pm$ 6.83 & 97.54 $\pm$ 1.71 & 49.73 $\pm$ 9.03 & 17.50 $\pm$ 20.46 & 0.460 $\pm$ 0.064 & \textbf{100.00 $\pm$ 0.00} & \textbf{79.61 $\pm$ 3.85} \\
Hierarchical & \textbf{98.05 $\pm$ 0.20} & 67.03 $\pm$ 8.25 & 98.02 $\pm$ 1.32 & 31.36 $\pm$ 13.67 & 40.71 $\pm$ 12.32 & 0.461 $\pm$ 0.057 & 98.66 $\pm$ 1.21 & 39.73 $\pm$ 2.66 \\
\textbf{Proposed} & \uline{97.57 $\pm$ 2.44} & \textbf{79.81 $\pm$ 3.27} & \uline{98.11 $\pm$ 1.24} & \textbf{72.83 $\pm$ 14.45} & \textbf{50.71 $\pm$ 6.10} & 0.467 $\pm$ 0.051 & \uline{99.85 $\pm$ 0.15} & 47.25 $\pm$ 0.87 \\
\end{tblr}
\end{table*}

%% file: chapters/results_ablation.tex
\subsection{Ablation Studies: Data Sources}
\label{sec:ablation_data}

To evaluate the contribution of each data source to the feature extraction process, an ablation study was conducted across the three study areas. We compared the performance of the proposed multi-branch architecture against two degraded configurations: (i) a single-branch model using only hyperspectral (HYP) data, and (ii) a dual-branch model combining HYP and panchromatic (PAN) data. To ensure a rigorously fair evaluation, both configurations were independently optimized from scratch, strictly following the same hyperparameter tuning protocol detailed for the standard architecture in Section \ref{sec:optimization}.

As reported in Table \ref{tab:ablation_branches_all_areas}, as expected, hyperspectral information provides the main discriminative contribution for both FT and TS classification across all study areas, confirming that spectral signatures represent the primary source of information for distinguishing \ac{FT}s and \ac{TS}s. Nevertheless, the integration of PAN provides a slight but consistent improvement in \ac{FT} classification, increasing \ac{OA} by 0.30, 0.52, and 0.37 percentage points in the Central-Eastern Alps, Central Apennines, and Southern Apennines, respectively. This consistent trend indicates that fine-scale spatial information contributes complementary structural information beyond that conveyed by the hyperspectral signatures alone. The inclusion of DEM shows a more pronounced and consistent effect on \ac{TS} classification, as the complete HYP+PAN+DEM configuration achieves the highest Comp. Acc. in all three study areas. Specifically, adding DEM to the HYP+PAN configuration improves Comp. Acc. by 2.50, 2.44, and 3.15 percentage points across the three areas, corresponding to an average gain of approximately 2.70 percentage points. The consistency and magnitude of these improvements indicate that topographic information provides relevant complementary cues for species discrimination, likely by accounting for environmental gradients associated with elevation and terrain morphology. For \ac{FT} classification, the differences among the configurations remain comparatively small: the complete architecture achieves the highest \ac{OA} in the Central-Eastern Alps, whereas the HYP+PAN configuration provides slightly higher \ac{OA} values in the two Apennine areas. Thus, the effect of DEM on \ac{FT} discrimination appears more dependent on the characteristics of the study area, whereas its contribution to \ac{TS} classification is substantially more systematic. Moreover, the HYP-only configuration exhibits the highest variability across folds in Comp. Acc. across all three study areas, suggesting that the integration of complementary data sources also contributes to more consistent TS classification performance. Overall, these results indicate that PAN and DEM provide complementary information to the HYP data, with distinct contributions to the two classification tasks: PAN consistently enhances \ac{FT} classification through the incorporation of fine-scale spatial information, while DEM provides a clearer benefit for \ac{TS} discrimination. In particular, the consistent increase in Comp. Acc. obtained with the complete configuration supports the inclusion of topographic information in the final \ac{TS} classification framework, while the systematic, albeit smaller, \ac{OA} gains obtained with PAN support the integration of spatial information within the proposed multi-branch architecture.

\definecolor{Gallery}{rgb}{0.950,0.950,0.950}
\begin{table}
\centering
\caption{Ablation study of the multi-branch feature extractor across the three study areas. Results quantify the incremental impact of HYP, PAN, and DEM integration on TS and FT classification performance in percentages (\%).}
\label{tab:ablation_branches_all_areas}

\begin{tblr}{
width = \linewidth,
colspec = {X[1.5,c] X[0.6,c] X[0.6,c] X[0.6,c] X[1.2,c] X[1.2,c]},
colsep = 2pt,
row{1,2} = {font=\bfseries, halign=c},
cell{1}{1} = {r=2}{c},
cell{1}{2} = {c=3}{},
cell{1}{5} = {r=2}{c},
cell{1}{6} = {r=2}{c},
cell{3}{1} = {r=3}{c,m},
cell{6}{1} = {r=3}{c,m},
cell{9}{1} = {r=3}{c,m},
hline{1,Z} = {0.2pt},
hline{3} = {0.5pt},
hline{2} = {2-4}{0.2pt},
hline{6,9} = {0.5pt},
cell{5}{5} = {Gallery,c},
cell{5}{6} = {Gallery,c},
cell{7}{5} = {Gallery,c},
cell{8}{6} = {Gallery,c},
cell{10}{5} = {Gallery,c},
cell{11}{6} = {Gallery,c},
}

Study Area & Data Sources & & & OA & Comp. Acc. \\
& HYP & PAN & DEM & & \\

Central-Eastern Alps
& \checkmark & & 
& $82.06 \pm 4.41$
& \uline{$78.06 \pm 2.84$} \\

& \checkmark & \checkmark &
& \uline{$82.36 \pm 3.73$}
& $77.76 \pm 1.60$ \\

& \checkmark & \checkmark & \checkmark
& \textbf{83.42 $\pm$ 4.63}
& \textbf{80.26 $\pm$ 2.67} \\

Central Apennines
& \checkmark & &
& $88.04 \pm 4.47$
& $72.15 \pm 2.44$ \\

& \checkmark & \checkmark &
& \textbf{88.56 $\pm$ 4.90}
& \uline{$72.15 \pm 1.38$} \\

& \checkmark & \checkmark & \checkmark
& \uline{$88.10 \pm 4.66$}
& \textbf{74.59 $\pm$ 0.98} \\

Southern Apennines
& \checkmark & &
& $97.85 \pm 1.31$
& \uline{$76.84 \pm 3.78$} \\

& \checkmark & \checkmark &
& \textbf{98.22 $\pm$ 1.10}
& $76.66 \pm 3.54$ \\

& \checkmark & \checkmark & \checkmark
& \uline{$98.16 \pm 1.15$}
& \textbf{79.81 $\pm$ 3.27} \\

\end{tblr}
\end{table}

\subsection{Ablation Studies: Components of the Proposed Framework}
\label{sec:ablation_components}

After validating the multi-branch feature extraction backbone across the three study areas, we analyzed the internal components of the joint \ac{FT}-\ac{TS} classifier. The objective is to isolate the individual contributions by comparing the complete proposed architecture against ablated configurations in which the cascade mechanism, the self-attention module, or the ecological constraints are selectively disabled. Consistent with the previous experiment, to ensure a rigorously fair comparison, each ablated network was independently optimized from scratch following the exact hyperparameter tuning protocol detailed in Section \ref{sec:optimization}. This analysis, evaluated in terms of Comp. Acc., clarifies how each architectural choice contributes to the overall consistency of the sub-pixel taxonomic predictions.

The results of this ablation study, presented in Table \ref{tab:ablation_classifier_all}, show that the Transformer provides the most consistent quantitative contribution across the three study areas, whereas the effects of the cascade classifier and the predefined ecological knowledge are clearly area-dependent. The largest and most systematic performance drop occurs when the Transformer is disabled, highlighting the critical role of the self-attention mechanism in explicitly modeling the dependencies between hierarchical levels. 

Furthermore, disabling the cascade classifier (i.e., predicting the sub-pixel species independently and simultaneously rather than sequentially) produces modest changes in the Central-Eastern Alps and Central Apennines, whereas in the Southern Apennines it yields the best Comp. Acc., but with substantially higher variability across folds, making the higher mean accuracy less indicative of a consistent improvement.

Finally, the ablation of the ecological knowledge reveals that the core representational power is provided by the Transformer architecture, while the explicit integration of domain expertise provides a consistent complementary refinement. Starting from a configuration using topology-constrained attention and the cascade but no predefined hierarchy, we evaluated the contribution of the ecological knowledge by comparing it with the complete configuration, in which the hierarchy matrix $\mathbf{H}$ is incorporated into the classifier and ecologically invalid transitions are penalized during inference. Across all three study areas, the inclusion of ecological knowledge consistently improves Comp. Acc. over the corresponding no-prior configuration, with gains of 0.57, 0.12, and 0.50 percentage points in the Central-Eastern Alps, Central Apennines, and Southern Apennines, respectively. While the Transformer provides the representational capacity to correlate labels and drives the overall performance, the full ecological masking explicitly guarantees that the final multi-label predictions are structurally admissible relative to the predicted \ac{FT} and the encoded hierarchy $\mathbf{H}$, provided that this hierarchy correctly represents the target classification system. The deterministic constraint does not by itself establish correspondence with the physical distribution of the forest biomes.

\begin{table}
\centering
\caption{Ablation study of the internal components of the joint \ac{FT}-\ac{TS} classifier across the three study areas. The Composite Accuracy (\%) is evaluated by selectively disabling the Ecological Knowledge (Ecol.\ Know.), the Transformer module (Transf.), and the Cascade Classifier (Cascade Cl.).}
\label{tab:ablation_classifier_all}

\begin{tblr}{
  width = \linewidth,
  colspec = {X[1.5,c] X[0.8,c] X[0.8,c] X[0.9,c] X[1.3,c]},
  colsep = 2pt,
  row{1} = {font=\bfseries, halign=c},
  cell{2}{1} = {r=4}{c,m},
  cell{6}{1} = {r=4}{c,m},
  cell{10}{1} = {r=4}{c,m},
  hline{1,Z} = {0.8pt},
  hline{2} = {0.4pt},
  hline{6,10} = {0.4pt},
  cell{2}{5} = {Gallery,c},
  cell{6}{5} = {Gallery,c},
  cell{12}{5} = {Gallery,c},
}

Study Area 
& Ecol. Know. 
& Transf. 
& Cascade Cl. 
& Comp. Acc. \\

Central-Eastern Alps
& \checkmark 
& \checkmark 
& \checkmark 
& \textbf{80.26 $\pm$ 2.67} \\

& $\times$ 
& \checkmark 
& \checkmark 
& $79.69 \pm 2.39$ \\

& \checkmark 
& \checkmark 
& $\times$ 
& \uline{$79.74 \pm 3.13$} \\

& $\times$ 
& $\times$ 
& \checkmark 
& $77.73 \pm 1.96$ \\

Central Apennines
& \checkmark 
& \checkmark 
& \checkmark 
& \textbf{74.59 $\pm$ 0.98} \\

& $\times$ 
& \checkmark 
& \checkmark 
& \uline{$74.47 \pm 1.27$} \\

& \checkmark 
& \checkmark 
& $\times$ 
& $74.18 \pm 2.11$ \\

& $\times$ 
& $\times$ 
& \checkmark 
& $72.66 \pm 1.37$ \\

Southern Apennines
& \checkmark 
& \checkmark 
& \checkmark 
& \uline{$79.81 \pm 3.27$} \\

& $\times$ 
& \checkmark 
& \checkmark 
& $79.31 \pm 3.98$ \\

& \checkmark 
& \checkmark 
& $\times$ 
& \textbf{82.29 $\pm$ 7.40} \\

& $\times$ 
& $\times$ 
& \checkmark 
& $78.80 \pm 3.86$ \\

\end{tblr}
\end{table}

%% file: chapters/conclusions.tex
In this study, we introduced a hierarchy-aware \ac{DL} framework specifically designed for regional-scale \ac{TS} mapping using PRISMA spaceborne hyperspectral data. By integrating spectral, spatial, and topographic information, the proposed multi-branch architecture addresses the challenge of sub-pixel heterogeneity in natural forest ecosystems. Our findings demonstrate that the structurally-masked Transformer effectively bridges the gap between broad \ac{FT} and fine-grained \ac{TS} compositions. A key contribution of this work is the implementation of a bidirectional learning strategy: while ecological constraints guide the species identification \textit{top-down}, the sub-pixel features simultaneously refine the \ac{FT} classification \textit{bottom-up}, enabling information exchange between the two tasks. The experimental evaluation across diverse Alpine and Apennine landscapes confirms that our method successfully resolves the relative dominance of co-occurring species, achieving the highest Comp. Acc. among the evaluated multi-label and ranking-based baselines. The ablation indicates that the Transformer provides the most consistent quantitative contribution across the different study areas. The cascade classifier contributes to stabilizing the predictions across data splits, consistently reducing fold-to-fold variability while maintaining competitive mean performance across the study areas. The predefined ecological knowledge provides a slight but consistent improvement across the study areas, while its main contribution lies in guaranteeing the admissibility of the predictions relative to the encoded hierarchy. However, this guarantee depends on the correctness and completeness of the predefined hierarchy and is enforced relative to the predicted FT. Furthermore, the ablation studies indicate that, as expected, the contribution of PAN and \acp{DEM} is generally limited compared with that of hyperspectral data. While these additional modalities lead to substantial improvements only in some cases, they provide complementary spatial and topographic information, with \acp{DEM} showing a more consistent benefit for discriminating spectrally similar species associated with different ecological conditions.

The results of this work open several promising directions for future research. The most significant development involves transitioning from ranked sub-pixel classification to estimation of species-specific fractional canopy cover, enabling a quantitative characterization of forest stand composition. To further refine these estimates, the integration of multitemporal hyperspectral acquisitions represents a natural progression to exploit phenological patterns, thereby improving the spectral separability of highly intermixed \acp{TS} throughout the vegetative cycle.

Finally, future efforts will focus on evaluating the interoperability and adaptability of the proposed hierarchical architecture with other spaceborne missions, such as EnMAP and the upcoming CHIME (Copernicus Hyperspectral Imaging Mission for the Environment). Although the present validation is limited to PRISMA acquisitions, the modular architecture could be adapted to sensors without a PAN band by omitting the PAN branch and reconfiguring the feature-fusion layer to combine hyperspectral and DEM features, thus enabling the assessment of cross-sensor generalization and potentially extending the framework to support consistent monitoring of forest biodiversity.

%% file: bib.bib
@article{dalponte2012tree,
  title={Tree species classification in boreal forests with hyperspectral data},
  author={Dalponte, Michele and {\O}rka, Hans Ole and Gobakken, Terje and Gianelle, Damiano and N{\ae}sset, Erik},
  journal={IEEE Trans. Geosci. Remote Sens.},
  volume={51},
  number={5},
  pages={2632--2645},
  year={2012},
  publisher={IEEE}
}

@article{ballanti2016tree,
  title={Tree species classification using hyperspectral imagery: A comparison of two classifiers},
  author={Ballanti, Laurel and Blesius, Leonhard and Hines, Ellen and Kruse, Bill},
  journal={Remote Sens.},
  volume={8},
  number={6},
  pages={445},
  year={2016},
  publisher={MDPI}
}

@article{hycza2018potential,
  title={Potential use of hyperspectral data to classify forest tree species},
  author={Hycza, Tomasz and Stere{\'n}czak, Krzysztof and Ba{\l}azy, Radomir},
  journal={N. Z. J. For. Sci.},
  volume={48},
  number={1},
  pages={18},
  year={2018},
  publisher={Springer}
}

@article{feret2012tree,
  title={Tree species discrimination in tropical forests using airborne imaging spectroscopy},
  author={F{\'e}ret, Jean-Baptiste and Asner, Gregory P},
  journal={IEEE Trans. Geosci. Remote Sens.},
  volume={51},
  number={1},
  pages={73--84},
  year={2012},
  publisher={IEEE}
}

@article{clark2005hyperspectral,
  title={Hyperspectral discrimination of tropical rain forest tree species at leaf to crown scales},
  author={Clark, Matthew L and Roberts, Dar A and Clark, David B},
  journal={Remote Sens. Environ.},
  volume={96},
  number={3-4},
  pages={375--398},
  year={2005},
  publisher={Elsevier}
}

@article{dalponte2008fusion,
  title={Fusion of hyperspectral and {LiDAR} remote sensing data for classification of complex forest areas},
  author={Dalponte, Michele and Bruzzone, Lorenzo and Gianelle, Damiano},
  journal={IEEE Trans. Geosci. Remote Sens.},
  volume={46},
  number={5},
  pages={1416--1427},
  year={2008},
  publisher={IEEE}
}

@article{cho2012mapping,
  title={Mapping tree species composition in {South African} savannas using an integrated airborne spectral and {LiDAR} system},
  author={Cho, Moses Azong and Mathieu, Renaud and Asner, Gregory P and Naidoo, Laven and Van Aardt, JAN and Ramoelo, Abel and Debba, Pravesh and Wessels, Konrad and Main, Russell and Smit, Izak PJ and others},
  journal={Remote Sens. Environ.},
  volume={125},
  pages={214--226},
  year={2012},
  publisher={Elsevier}
}

@article{vangi2021new,
  title={The new hyperspectral satellite {PRISMA}: Imagery for forest types discrimination},
  author={Vangi, Elia and D’Amico, Giovanni and Francini, Saverio and Giannetti, Francesca and Lasserre, Bruno and Marchetti, Marco and Chirici, Gherardo},
  journal={Sensors},
  volume={21},
  number={4},
  pages={1182},
  year={2021},
  publisher={MDPI}
}

@article{laurin2025monitoring,
  title={Monitoring habitat diversity with {PRISMA} hyperspectral and lidar-derived data in {Natura 2000} sites: Case study from a {Mediterranean} forest},
  author={Laurin, Gaia Vaglio and Zabeo, Chiara and Giuliarelli, Diego and Tesfamariam, Birhane Gebrehiwot and Cotrina-Sanchez, Alexander and Valentini, Riccardo and Tufail, Basil and Ventura, Bartolomeo and Calfapietra, Carlo and Barbati, Anna},
  journal={Ecol. Indic.},
  volume={172},
  pages={113254},
  year={2025},
  publisher={Elsevier}
}

@article{delogu2024leveraging,
  title={Leveraging the Potential of {PRISMA} Hyperspectral Data for Forest Tree Species Classification: A Case Study in {Southern Italy}},
  author={Delogu, Gabriele and Perretta, Miriam and Caputi, Eros and Patriarca, Alessio and Funsten, Cassandra Carroll and Recanatesi, Fabio and Ripa, Maria Nicolina and Boccia, Lorenzo},
  journal={Remote Sens.},
  volume={16},
  number={24},
  pages={4788},
  year={2024},
  publisher={MDPI}
}

@article{fassnacht2016review,
  title={Review of studies on tree species classification from remotely sensed data},
  author={Fassnacht, Fabian Ewald and Latifi, Hooman and Stere{\'n}czak, Krzysztof and Modzelewska, Aneta Pins e, Michael and Waser, Lars T and Straub, Christoph and Ghosh, Aniruddha},
  journal={Remote Sens. Environ.},
  volume={186},
  pages={64--87},
  year={2016},
  publisher={Elsevier}
}

@article{paoletti2019deep,
  title={Deep learning classifiers for hyperspectral imaging: A review},
  author={Paoletti, Mercedes Eugenia and Haut, Juan Mario and Plaza, Javier and Plaza, Antonio},
  journal={ISPRS J. Photogramm. Remote Sens.},
  volume={158},
  pages={279--317},
  year={2019},
  publisher={Elsevier}
}

@article{hong2022spectralformer,
  title={Spectralformer: Rethinking hyperspectral image classification with transformers},
  author={Hong, Danfeng and Han, Zhu and Yao, Jing and Gao, Lianru and Zhang, Bing and Plaza, Antonio and Chanussot, Jocelyn},
  journal={IEEE Trans. Geosci. Remote Sens.},
  year={2022},
  volume={60},
  pages={1-15},
  note = {DOI: 10.1109/TGRS.2021.3130716}
}

@article{zhang2020three,
  title={Three-dimensional convolutional neural network model for tree species classification using airborne hyperspectral images},
  author={Zhang, Bin and Zhao, Lin and Zhang, Xiaoli},
  journal={Remote Sens. Environ.},
  volume={247},
  pages={111938},
  year={2020},
  publisher={Elsevier}
}

@article{xi2019mapping,
  title={Mapping tree species composition using {OHS-1} hyperspectral data and deep learning algorithms in {Changbai} mountains, {Northeast China}},
  author={Xi, Yanbiao and Ren, Chunying and Wang, Zongming and Wei, Shiqing and Bai, Jialing and Zhang, Bai and Xiang, Hengxing and Chen, Lin},
  journal={Forests},
  volume={10},
  number={9},
  pages={818},
  year={2019},
  publisher={MDPI}
}

@article{nezami2020tree,
  title={Tree species classification of drone hyperspectral and {RGB} imagery with deep learning convolutional neural networks},
  author={Nezami, Somayeh and Khoramshahi, Ehsan and Nevalainen, Olli and P{\"o}l{\"o}nen, Ilkka and Honkavaara, Eija},
  journal={Remote Sens.},
  volume={12},
  number={7},
  pages={1070},
  year={2020},
  publisher={MDPI}
}

@article{sothe2020evaluating,
  title={Evaluating a convolutional neural network for feature extraction and tree species classification using {UAV}-hyperspectral images},
  author={Sothe, C. and La Rosa, L. E. C. and De Almeida, C. M. and Gonsamo, A. and Schimalski, M. B. and Castro, J. D. B. and Feitosa, R. Q. and Dalponte, M. and Lima, C. L. and Liesenberg, V. and others},
  journal={ISPRS Ann. Photogramm. Remote Sens. Spatial Inf. Sci.},
  volume={3},
  pages={193--199},
  year={2020}
}

@article{fassnacht2014comparison,
  title={Comparison of feature reduction algorithms for classifying tree species with hyperspectral data on three central {European} test sites},
  author={Fassnacht, Fabian E. and Neumann, Carsten and Foerster, Michael and Buddenbaum, Henning and Ghosh, Aniruddha and Clasen, Anne and Joshi, Pawan Kumar and Koch, Barbara},
  journal={IEEE J. Sel. Topics Appl. Earth Observ. Remote Sens.},
  volume={7},
  number={6},
  pages={2547--2561},
  year={2014},
  publisher={IEEE}
}

@article{fricker2019convolutional,
  title={A convolutional neural network classifier identifies tree species in mixed-conifer forest from hyperspectral imagery},
  author={Fricker, Geoffrey A. and Ventura, Jonathan D. and Wolf, Jeffrey A. and North, Malcolm P. and Davis, Frank W. and Franklin, Janet},
  journal={Remote Sens.},
  volume={11},
  number={19},
  pages={2326},
  year={2019},
  publisher={MDPI}
}

@article{breidenbach2021national,
  title={National mapping and estimation of forest area by dominant tree species using {Sentinel-2} data},
  author={Breidenbach, Johannes and Waser, Lars T. and Debella-Gilo, Misganu and Schumacher, Johannes and Rahlf, Johannes and Hauglin, Marius and Puliti, Stefano and Astrup, Rasmus},
  journal={Can. J. For. Res.},
  volume={51},
  number={3},
  pages={365--379},
  year={2021}
}

@article{hoscilo2019mapping,
  title={Mapping forest type and tree species on a regional scale using multi-temporal {Sentinel-2} data},
  author={Ho{\'s}ci{\l}o, Agata and Lewandowska, Aneta},
  journal={Remote Sens.},
  volume={11},
  number={8},
  pages={929},
  year={2019},
  publisher={MDPI}
}

@article{vanguri2024mapping,
  title={Mapping forest tree species and its biodiversity using {EnMAP} hyperspectral data along with {Sentinel-2} temporal data},
  author={Vanguri, Rajesh and Laneve, Giovanni and Ho{\'s}ci{\l}o, Agata},
  journal={Ecol. Indic.},
  volume={167},
  pages={112671},
  year={2024},
  publisher={Elsevier}
}

@article{zhang2017spectral,
  title={Spectral-spatial classification of hyperspectral imagery using a dual-channel convolutional neural network},
  author={Zhang, Haokui and Li, Ying and Zhang, Yuzhu and Shen, Qiang},
  journal={Remote Sens. Lett.},
  volume={8},
  number={5},
  pages={438--447},
  year={2017},
  publisher={Taylor \& Francis}
}

@article{ma2018hyperspectral,
  title={Hyperspectral image classification based on deep deconvolution network with skip architecture},
  author={Ma, Xiaorui and Fu, Anyan and Wang, Jie and Wang, Hongyu and Yin, Baocai},
  journal={IEEE Trans. Geosci. Remote Sens.},
  volume={56},
  number={8},
  pages={4781--4791},
  year={2018},
  publisher={IEEE}
}

@article{chen2020mapping,
  title={Mapping multiple tree species classes using a hierarchical procedure with optimized node variables},
  author={Chen, Yaoliang and Zhao, Shuai and Xie, Zhuli and Lu, Dengsheng and Chen, Erxue},
  journal={GIScience Remote Sens.},
  volume={57},
  number={4},
  pages={526--542},
  year={2020},
  publisher={Taylor \& Francis}
}

@article{scheffler2017arosics,
  title={{AROSICS}: An automated and robust open-source image co-registration software for multi-sensor satellite data},
  author={Scheffler, Daniel and Hollstein, Andr{\'e} and Diedrich, Hannes and Segl, Karl and Hostert, Patrick},
  journal={Remote Sens.},
  volume={9},
  number={7},
  pages={676},
  year={2017},
  publisher={MDPI}
}

@article{de2024improving,
  title={Improving {PRISMA} hyperspectral spatial resolution and geolocation by using {Sentinel-2}},
  author={De Luca, Giandomenico and Carotenuto, Federico and Genesio, Lorenzo and Pepe, Monica and Toscano, Piero and Boschetti, Mirco and Miglietta, Franco and Gioli, Beniamino},
  journal={ISPRS J. Photogramm. Remote Sens.},
  volume={215},
  pages={112--135},
  year={2024},
  publisher={Elsevier}
}

@article{gasparini2025open,
  title={Open Georeferenced Field Data on Forest Types and Species for Biodiversity Assessment},
  author={Gasparini, Patrizia and Di Cosmo, Lucio and Floris, Antonio and Murgia, Federica and Rizzo, Maria},
  journal={Data},
  volume={10},
  number={3},
  pages={30},
  year={2025},
  publisher={MDPI}
}

@article{upadhyay2018hyperspectral,
  title={Hyperspectral remote sensing of forests: Technological advancements, opportunities and challenges},
  author={Upadhyay, Vipin and Kumar, Amit},
  journal={Earth Sci. Informat.},
  volume={11},
  number={4},
  pages={487--524},
  year={2018},
  publisher={Springer}
}

@inproceedings{candela2016prisma,
  title={The {PRISMA} mission},
  author={Candela, Laura and Formaro, Roberto and Guarini, Rocchina and Loizzo, Rosa and Longo, Francesco and Varacalli, Giancarlo},
  booktitle={Proc. IEEE Int. Geosci. Remote Sens. Symp. (IGARSS)},
  pages={253--256},
  year={2016}
}

@article{guanter2015enmap,
  title={The {EnMAP} spaceborne imaging spectroscopy mission for {Earth} observation},
  author={Guanter, Luis and Kaufmann, Hermann and Segl, Karl and Foerster, Saskia and Rogass, Christian and Chabrillat, Sabine and Kuester, Theres and Hollstein, Andr{\'e} and Rossner, Godela and Chlebek, Christian and others},
  journal={Remote Sens.},
  volume={7},
  number={7},
  pages={8830--8857},
  year={2015},
  publisher={MDPI}
}

@inproceedings{tan2021efficientnetv2,
  title={{EfficientNetV2}: Smaller models and faster training},
  author={Tan, Mingxing and Le, Quoc},
  booktitle={Proc. Int. Conf. Mach. Learn. (ICML)},
  pages={10096--10106},
  year={2021}
}

@incollection{gasparini2022definitions,
  title={Definitions and Sampling Design},
  author={Gasparini, Patrizia and Floris, Antonio},
  booktitle={Italian National Forest Inventory—Methods and Results},
  pages={17--48},
  year={2022},
  publisher={Springer}
}

@article{tarquini2023tinitaly,
  title={{TINITALY}, a digital elevation model of {Italy} with a 10 meters cell size},
  author={Tarquini, Simone and Isola, Ilaria and Favalli, Massimiliano and Battistini, Alessandro and Dotta, Giulia},
  journal={Ist. Naz. Geofis. Vulcanol. (INGV)},
  volume={10},
  year={2023}
}

@article{bioucas2012hyperspectral,
  title={Hyperspectral unmixing overview: Geometrical, statistical, and sparse regression-based approaches},
  author={Bioucas-Dias, Jos{\'e} M. and Plaza, Antonio and Dobigeon, Nicolas and Parente, Mario and Du, Qian and Gader, Paul and Chanussot, Jocelyn},
  journal={IEEE J. Sel. Topics Appl. Earth Observ. Remote Sens.},
  volume={5},
  number={2},
  pages={354--379},
  year={2012},
  publisher={IEEE}
}

@article{silla2011survey,
  title={A survey of hierarchical classification across different application domains},
  author={Silla Jr, Carlos N. and Freitas, Alex A.},
  journal={Data Min. Knowl. Discov.},
  volume={22},
  number={1},
  pages={31--72},
  year={2011},
  publisher={Springer}
}

@article{hu2015deep,
  title={Deep convolutional neural networks for hyperspectral image classification},
  author={Hu, Wei and Huang, Yangyu and Wei, Li and Zhang, Fan and Li, Hengchao},
  journal={J. Sensors},
  volume={2015},
  year={2015},
  publisher={Wiley}
}

@article{bergstra2011algorithms,
  title={Algorithms for hyper-parameter optimization},
  author={Bergstra, James and Bardenet, R{\'e}mi and Bengio, Yoshua and K{\'e}gl, Bal{\'a}zs},
  journal={Adv. Neural Inf. Process. Syst. (NeurIPS)},
  volume={24},
  year={2011}
}

@article{karalas2016land,
  title={Land classification using remotely sensed data: Going multilabel},
  author={Karalas, Konstantinos and Tsagkatakis, Grigorios and Zervakis, Michael and Tsakalides, Panagiotis},
  journal={IEEE Trans. Geosci. Remote Sens.},
  volume={54},
  number={6},
  pages={3548--3563},
  year={2016},
  publisher={IEEE}
}

@article{sumbul2020deep,
  title={A deep multi-attention driven approach for multi-label remote sensing image classification},
  author={Sumbul, Gencer and Dem{\.I}r, Beg{\"u}m},
  journal={IEEE Access},
  volume={8},
  pages={95934--95946},
  year={2020},
  publisher={IEEE}
}

@article{zhang2018binary,
  title={Binary relevance for multi-label learning: an overview},
  author={Zhang, Min-Ling and Li, Yu-Kun and Liu, Xu-Ying and Geng, Xin},
  journal={Front. Comput. Sci.},
  volume={12},
  number={2},
  pages={191--202},
  year={2018},
  publisher={Springer}
}

@inproceedings{cao2007learning,
  title={Learning to rank: from pairwise approach to listwise approach},
  author={Cao, Zhe and Qin, Tao and Liu, Tie-Yan and Tsai, Ming-Feng and Li, Hang},
  booktitle={Proc. 24th Int. Conf. Mach. Learn. (ICML)},
  pages={129--136},
  year={2007}
}

@article{sen2022hierarchical,
  title={A hierarchical approach to remote sensing scene classification},
  author={Sen, Ozlem and Keles, Hacer Yalim},
  journal={PFG--J. Photogramm. Remote Sens. Geoinf. Sci.},
  volume={90},
  number={2},
  pages={161--175},
  year={2022},
  publisher={Springer}
}

@article{liu2026switcher,
  title={{Switcher-HNet}: A switchable hierarchical network for tree species classification from forest stand to individual tree tasks},
  author={Liu, Shengheng and Li, Xiang Dean, Zhen and Zhao, Yinghui},
  journal={ISPRS J. Photogramm. Remote Sens.},
  volume={231},
  pages={329--344},
  year={2026},
  publisher={Elsevier}
}

@article{qin2010general,
  title={A general approximation framework for direct optimization of information retrieval measures},
  author={Qin, Tao and Liu, Tie-Yan and Li, Hang},
  journal={Inf. Retr.},
  volume={13},
  number={4},
  pages={375--397},
  year={2010},
  publisher={Springer}
}

@article{tsoumakas2007multi,
  title={Multi-label classification: An overview},
  author={Tsoumakas, Grigorios and Katakis, Ioannis},
  journal={Int. J. Data Warehousing Mining},
  volume={3},
  number={3},
  pages={1--13},
  year={2007},
  publisher={IGI Global}
}

@misc{copernicus2021,
  title={Copernicus Forest Type 2018 Europe Map},
  howpublished={European Union's Copernicus Land Monitoring Service information},
  year={2018},
  doi={10.2909/77873ff3-4edf-48d4-94cd-c5b7b61da29e}
}

@incollection{rizzo2022land,
  title={Land Use and Land Cover Photointerpretation},
  author={Rizzo, Maria and Gasparini, Patrizia},
  booktitle={I. N. F. I.: Methods and Results of the Third Survey},
  pages={49--66},
  year={2022},
  publisher={Springer}
}

@article{zhong2024review,
  title={A review: Tree species classification based on remote sensing data and classic deep learning-based methods},
  author={Zhong, Lihui and Dai, Zhengquan and Fang, Panfei and Cao, Yong and Wang, Leiguang},
  journal={Forests},
  volume={15},
  number={5},
  pages={852},
  year={2024},
  publisher={MDPI}
}

@article{chaity2024exploring,
  title={Exploring the limits of species identification via a convolutional neural network in a complex forest scene through simulated imaging spectroscopy},
  author={Chaity, Manisha Das and van Aardt, Jan},
  journal={Remote Sens.},
  volume={16},
  number={3},
  pages={498},
  year={2024},
  publisher={MDPI}
}

@inproceedings{polonen2018tree,
  title={Tree species identification using {3D} spectral data and {3D} convolutional neural network},
  author={P{\"o}l{\"o}nen, Ilkka and Annala, Leevi and Rahkonen, Samuli and Nevalainen, Olli and Honkavaara, Eija and Tuominen, Sakari and Viljanen, Niko and Hakala, Teemu},
  booktitle={Proc. WHISPERS},
  pages={1--5},
  year={2018},
  organization={IEEE}
}

@inproceedings{forgavc2023contribution,
  title={Contribution to pixel-based hyperspectral classification of tree species by {1D-CNN}},
  author={Forg{\'a}{\v{c}}, Radoslav and Badidov{\'a}, Bianca and O{\v{c}}kay, Milo{\v{s}} and Krammer, Peter and Javurek, Martin and Bil'ansk{\'a}, Michaela},
  booktitle={Proc. Commun. Inf. Technol. (KIT)},
  pages={1--5},
  year={2023},
  organization={IEEE}
}

@article{braham2025spectralearth,
  title={{Spectralearth}: Training hyperspectral foundation models at scale},
  author={Braham, Nassim Ait Ali and Albrecht, Conrad M and Mairal, Julien and Chanussot, Jocelyn and Wang, Yi and Zhu, Xiao Xiang},
  journal={IEEE J. Sel. Topics Appl. Earth Observ. Remote Sens.},
  year={2025},
  publisher={IEEE}
}

@article{wang2025hypersigma,
  title={{HyperSIGMA}: Hyperspectral intelligence comprehension foundation model},
  author={Wang, Di and Hu, Meiqi and Jin, Yao and Miao, Yuchun and Yang, Jiaqi and Xu, Yichu and Qin, Xiaolei and Ma, Jiaqi and Sun, Lingyu and Li, Chenxing and others},
  journal={IEEE Trans. Pattern Anal. Mach. Intell.},
  year={2025},
  publisher={IEEE}
}

@article{somers2014tree,
  title={Tree species mapping in tropical forests using multi-temporal imaging spectroscopy: Wavelength adaptive spectral mixture analysis},
  author={Somers, Ben and Asner, Gregory P},
  journal={Int. J. Appl. Earth Observ. Geoinf.},
  volume={31},
  pages={57--66},
  year={2014},
  publisher={Elsevier}
}

@article{stagakis2016estimating,
  title={Estimating forest species abundance through linear unmixing of {CHRIS/PROBA} imagery},
  author={Stagakis, Stavros and Vanikiotis, Theofilos and Sykioti, Olga},
  journal={ISPRS J. Photogramm. Remote Sens.},
  volume={119},
  pages={79--89},
  year={2016},
  publisher={Elsevier}
}

@article{quan2023tree,
  title={Tree species classification in a typical natural secondary forest using {UAV}-borne {LiDAR} and hyperspectral data},
  author={Quan, Ying and Li, Mingze and Hao, Yuanshuo and Liu, Jianyang and Wang, Bin},
  journal={GIScience Remote Sens.},
  volume={60},
  number={1},
  pages={2171706},
  year={2023},
  publisher={Taylor \& Francis}
}

@article{ma2024deep,
  title={A deep-learning-based tree species classification for natural secondary forests using unmanned aerial vehicle hyperspectral images and {LiDAR}},
  author={Ma, Ye and Zhao, Yuting and Im, Jungho and Zhao, Yinghui and Zhen, Zhen},
  journal={Ecol. Indic.},
  volume={159},
  pages={111608},
  year={2024},
  publisher={Elsevier}
}

@article{yu2025forest,
  title={Forest Tree Species Classification Based on Deep Ensemble Learning by Fusing High-Resolution, Multitemporal, and Hyperspectral Multisource Remote Sensing Data},
  author={Yu, Dengli and Tu, Lilin and Wei, Ziqing and Zhu, Fuyao and Yu, Chengjun and Wang, Denghong and Li, Jiayi and Huang, Xin},
  journal={IEEE Geosci. Remote Sens. Lett.},
  volume={23},
  pages={1--5},
  year={2025},
  publisher={IEEE}
}

@article{pan2024hyperspectral,
  title={Hyperspectral Image Classification with Mamba},
  author={Pan, Zhaojie and Li, Chenyu and Plaza, Antonio and Chanussot, Jocelyn and Hong, Danfeng},
  journal={IEEE Trans. Geosci. Remote Sens.}, 
  year={2024},
  note = {DOI: 10.1109/TGRS.2024.3521411}
}

@article{kluczek2023mountain,
  title={Mountain tree species mapping using Sentinel-2, PlanetScope, and airborne HySpex hyperspectral imagery},
  author={Kluczek, Marcin and Zagajewski, Bogdan and Zwijacz-Kozica, Tomasz},
  journal={Remote Sensing},
  volume={15},
  number={3},
  pages={844},
  year={2023},
  publisher={MDPI}
}

@article{vorovencii2024assessing,
  title={Assessing various scenarios of multitemporal Sentinel-2 imagery, topographic data, texture features, and machine learning algorithms for tree species identification},
  author={Vorovencii, Iosif},
  journal={IEEE Journal of Selected Topics in Applied Earth Observations and Remote Sensing},
  volume={17},
  pages={15373--15392},
  year={2024},
  publisher={IEEE}
}

@article{perretta2024testing,
  title={Testing the impact of pansharpening using PRISMA hyperspectral data: A case study classifying urban trees in Naples, Italy},
  author={Perretta, Miriam and Delogu, Gabriele and Funsten, Cassandra and Patriarca, Alessio and Caputi, Eros and Boccia, Lorenzo},
  journal={Remote Sensing},
  volume={16},
  number={19},
  pages={3730},
  year={2024},
  publisher={MDPI}
}

@article{wan2020hyperspectral,
  title={Hyperspectral image classification with context-aware dynamic graph convolutional network},
  author={Wan, Sheng and Gong, Chen and Zhong, Ping and Pan, Shirui and Li, Guangyu and Yang, Jian},
  journal={IEEE Transactions on Geoscience and Remote Sensing},
  volume={59},
  number={1},
  pages={597--612},
  year={2020},
  publisher={IEEE}
}
